\documentclass[natbib,openany]{sn-jnl}
\usepackage{natbib}

\usepackage{graphics}%
\usepackage{multirow}%
\usepackage{amsmath,amssymb,amsfonts}%
\usepackage{scalerel,stackengine}
\usepackage{amsthm}%
\usepackage{mathrsfs}%
\usepackage[title]{appendix}%
\usepackage[dvipsames]{xcolor}%

\definecolor{myred}{RGB}{158, 15, 20}
\definecolor{mypink}{RGB}{255, 128, 128}

\usepackage{textcomp}%
\usepackage{manyfoot}%
\usepackage{booktabs}%
\usepackage{algorithm}%
\usepackage{algorithmicx}%
\usepackage{algpseudocode}%
\usepackage{listings}%

\definecolor{code}{HTML}{f7f7f7} 

\newfloat{lstfloat}{htbp}{lop}
\floatname{lstfloat}{Listing}

\usepackage{changepage}

\usepackage{apacdoc} 

\usepackage{tikz}

\usepackage{rotating}
\usepackage{float}
\usepackage{dcolumn}       
\usepackage{tablefootnote}
\usepackage{array}
\usepackage{adjustbox}

\usepackage{lscape}

\newcolumntype{R}[2]{%
	>{\adjustbox{angle=#1,lap=\width-(#2)}\bgroup}%
	l%
	<{\egroup}%
}

\usepackage{mathtools}

\usepackage{arydshln}
\usepackage{boldline}
\usepackage{colortbl}
\usepackage{tikz}
\usetikzlibrary{arrows.meta,
	calc, chains,
	decorations.pathreplacing,%
	calligraphy,
	shapes,
	shapes.geometric, arrows, matrix, positioning, fit, calc}
\definecolor{blue1}{RGB}{84,141,212}
\definecolor{blue2}{RGB}{142,180,227}
\definecolor{yellow1}{RGB}{255,229,153}
\definecolor{orange1}{RGB}{255,153,0}
\definecolor{gray1}{RGB}{127,127,127}
\definecolor{gray2}{RGB}{217,217,217}
\definecolor{DarkGreen}{RGB}{22,138,66}

\usepackage{bm}
\usepackage{dsfont}
\usepackage{pdflscape}
\usepackage{ltxtable, tabularx}
\usepackage{geometry}

\newcounter{alphasect}
\def\alphainsection{0}

\let\oldsection=\section
\def\section{%
	\ifnum\alphainsection=1%
	\addtocounter{alphasect}{1}
	\fi%
	\oldsection}%

\renewcommand\thesection{%
	\ifnum\alphainsection=1%
	\Alph{alphasect}
	\else%
	\arabic{section}
	\fi%
}%

\usepackage{bbm}
\usepackage{float}

\floatstyle{plaintop}
\restylefloat{table}

\usepackage{subcaption}
\usepackage{caption}

\usepackage{stackengine}

\theoremstyle{thmstyleone}%
\theoremstyle{thmstyletwo}%

\theoremstyle{thmstylethree}%

\begin{document}
	
	
	\title[Article Title]{Comparison of the Cox proportional hazards model and Random Survival Forest algorithm for predicting patient-specific survival probabilities in clinical trial data}
	
	
	\author*[1]{\fnm{Ricarda} \sur{Graf}}\email{ricarda.graf@math.uni-augsburg.de}
	
	\author[2]{\fnm{Susan} \sur{Todd}}\email{s.c.todd@reading.ac.uk}
	
	\author[2]{\fnm{M. Fazil} \sur{Baksh}}\email{m.f.baksh@reading.ac.uk}

	\affil[1]{\orgdiv{Institute of Mathematics}, \orgname{University of Augsburg}, \orgaddress{\city{Augsburg}, \postcode{86159},  \country{Germany}}}
	
	\affil[2]{\orgdiv{Department of Mathematics and Statistics}, \orgname{University of Reading}, \orgaddress{\city{Reading}, \postcode{RG6 6AX}, \country{UK}}}

	
	\abstract{
		
		
		The Cox proportional hazards model is often used to analyze data from Randomized Controlled Trials (RCT) with time-to-event outcomes. Random survival forest (RSF) is a machine-learning algorithm known for its high predictive performance.  We conduct a comprehensive neutral comparison study to compare the performance of Cox regression and RSF in various simulation scenarios based on two reference datasets from RCTs. The motivation is to identify settings in which one method is preferable over the other when comparing different aspects of performance using measures according to the TRIPOD (Transparent Reporting of a multivariable prediction model for Individual Prognosis or Diagnosis) recommendations.\\
		Our results show that conclusions solely based on the $C$ index, a performance measure that has been predominantly used in previous studies comparing predictive accuracy of the Cox-PH and RSF model based on real-world observational time-to-event data and that has been criticized by methodologists, may not be generalizable to other aspects of predictive performance. We found that measures of overall performance may generally give more reasonable results, and that the standard log-rank splitting rule used for the RSF may be outperformed by alternative splitting rules, in particular in nonproportional hazards settings.   In our simulations, performance of the RSF suffers less in data with treatment-covariate interactions compared to data where these are absent. Performance of the Cox-PH model is affected by the violation of the proportional hazards assumption.

		}
		

		
		
		\maketitle
		

		\section{Introduction}\label{sec1}
		
		Prognostic prediction models 
		are used to estimate an individual's probability based on multiple risk factors that a disease or outcome will occur in a specific period of time. They are most often used at time of diagnosis or start of treatment to support physicians in early detection, diagnosis, treatment decision, and prognosis, and to inform patients about their risks \citep{Moons2012}. They are applied in the medical field in general, and in particular in the field of cancer  treatment and research, the field of diabetes, and the cardiovascular field  \citep{Moons2012, Goldstein2016}. Clinical decision tools such as ``ClinicalPath'' \citep{Elsevier2022} for cancer treatment or the Framingham Risk Score \citep{Wilson1998} for coronary heart disease, are examples of prognostic prediction models.\\
		Cox regression \citep{Cox1972} is most widely used for developing prognostic models in medical time-to-event data \citep{Goldstein2016, Collins2011, Collins2014, Mahar2017, Mallett2010, Steyerberg2013, Wynants2019, Phung2019, Hueting20222}.  
		It provides estimates of the hazard ratios for each explanatory variable. In the context of clinical trials, the treatment effect hazard ratio is of particular interest. 
		As a semi-parametric model, it is assumed that at least 10 events have to be observed per predictor variable included in the Cox model to obtain reasonable results \citep{Peduzzi1995, Vittinghoff2006}, so it cannot be used in high-dimensional settings with a large number of potential predictor variables compared to the number of individuals. During model development, researchers often have to decide on a fraction of available predictors to be included in the final model \citep{Moons2012}.
		Cox regression requires  choice of terms to include in a model,  including possible (higher order) interaction terms and variable transformations in case of nonlinear relationships of continuous covariates with the survival outcome. Moreover, it makes the assumption of proportional hazards which means that it assumes the hazard ratio of any two patients  to be constant over the period of follow-up.
		In cases where the new treatment only shows an advantage at an early or later stage, respectively, interpretation of its results may not be meaningful. Especially in long-term studies, this assumption may be violated \citep{Hilsenbeck1998}.
		On the other hand, the Cox model provides corresponding measures of uncertainty (confidence intervals for the hazard ratios), which generally form the basis for clinical decision making, is easy to use and has short computational times. When using the  Cox model for predictions, the specification of a baseline survival distribution is required \citep{Therneau2000}.\\
		In comparison, the Random survival forest (RSF) algorithm  \citep{Ishwaran2008} is a nonparametric machine-learning approach. It is suitable for the same variable types as the Cox model, i.e. continuous right-censored survival time outcomes  and continuous as well as categorical predictor variables. In contrast, it does not require an explicit specification of a model but is able to detect and incorporate even complex interactions between the covariates and the survival outcome as well as nonlinear relationships. It is also suitable for a large number of covariates, although it is advisable not to include variables that are already suspected not to be meaningful in order  to not unnecessarily increase computational complexity. It also seems suitable for dependent censoring \citep{Zhou2014}. Moreover, it does not require the proportional hazard assumption. However, since the RSF does not make parametric assumptions regarding the data, it also does not provide uncertainty measures such as confidence intervals for its estimates.
		Machine-learning methods such as random forests have proven to increase predictive accuracy in prognostic studies \citep{Murmu2024}, especially 
		in high-dimensional data such as genetic, protein, or imaging biomarkers \citep{Cohen2018, Zhang2020, Kawakami2019, Lin2022, Ruyssinck2016}. For example, a prognostic model for glioblastoma widely used for more than two decades and most recently adapted to incorporate further relevant covariates \citep{Bell2017}, is based on a survival tree method. The RSF may help predicting patient outcomes and survival rates more accurately. Therefore the current work aims to explore the potential application of the RSF to data from randomized controlled trials (RCT) by comparing its predictive performance to the Cox proportional hazards (Cox-PH) model.  \\
		Previous studies compared the performance of Cox regression and RSF in observational clinical data, more specifically real-world datasets \citep[e.g.][]{Guo2023, Sarica2023, Chowdhury2023, Moncada2021, Farhadian2021, Miao2015, Spooner2020,Qiu2020,  Kim2019, Datema2012,  Omurlu2009, Du2020}. The predominantly  used performance measure in these studies is  Harrell's $C$ index \citep{Harrell1982, Harrell1996}, a rank-based measure of discrimination. 
		Very rarely, calibration, or overall performance are assessed. Only the study by Du et al. (\citeyear{Du2020}) considered all three recommended types of performance measures. In their systematic review and meta-analysis of 52 studies predicting hypertension, Chowdhury et al. (\citeyear{Chowdhury2022}) compared the performance of regression approaches (including  Cox regression) and various machine-learning methods (RSF was not applied in any of the studies). These authors too found that performance comparison based on the  $C$ index was common in contrast to comparisons based on calibration. Most of the above mentioned studies aiming to compare the performance of the Cox and RSF model stated an at least slightly better performance of the RSF model with respect to the $C$ index.  \\
		According to our literature search, only one study previously compared the two  approaches
		based on data simulations  \citep{Baralou2022}, for which reference data is taken from an observational study.. 
		Moreover, their comparison is not only based on the default log-rank splitting rule for the RSF, but includes two further splitting rules. In addition to a measure of discrimination (time-dependent area under the curve, AUC), they also use a measure of overall performance (Integrated Brier score, IBS \citep{Graf1999}). 		
		Most notably, they found that the RSF outperformed the Cox-PH model in scenarios with lower censoring rates in the presence of covariate interactions. However, they do not examine the performance for data from randomized controlled trials (including factors specific to RCTs such as different sizes of treatment effect, the absence/presence of treatment-covariate interactions, and smaller sample sizes less than 500), the influence of violation of the proportional hazard assumption, other splitting rules available for the RSF, and measures of calibration.  \\
		The  TRIPOD (Transparent Reporting of a multivariable prediction model for Individual Prognosis or Diagnosis) recommendations \citep{Moons2015} state that prognostic models should be compared with respect to discrimination (e.g. Harrell's $C$ index, time-dependent AUC for time-to-event data), calibration, and overall performance (e.g. Integrated Brier score). These three aspects of model performance are also described in Steyerberg et al. (\citeyear{Steyerberg2010}) and McLernon et al. (\citeyear{McLernon2023}), for example. Here, Harrell's $C$ index, calibration curves, and the Integrated Brier score will be used as performance measures, which are described in Section \ref{perf_meas}. \\
		Responsible integration of machine learning algorithms in any step of a clinical trial may help overcome some of the challenges in its design, conduct, and analysis, e.g. with respect to patient recruitment, or the planning of treatment interventions \citep{Miller2023, Weissler2021}. Evidence is needed where machine learning algorithms can be applied in order to gain an advantage such as more precise predictions free from parametric model assumptions.  \\ 
		Our simulation study may be the first one comparing the performance of the Cox-PH and RSF model for clinical trial settings. The aim is to evaluate the predictive accuracy of both methods, the Cox regression model and the RSF algorithm, in predicting patient-specific survival probabilities in right-censored clinical trial data. Two possible scenarios are considered, where  treatment-covariate interactions  in the data are either absent or present. For this purpose, two publicly available clinical trial datasets \citep{UMASS, Byar1980} without and with known treatment-covariate interactions serve as a reference for data simulations. 
		In contrast to previous studies, the performance of all six  RSF splitting rules (currently available in the most commonly used \texttt{R} packages \texttt{randomForestSRC} \citep{Ishwaran2023} and \texttt{ranger} \citep{Wright2023} is compared, and  evaluation is based on  measures of discrimination, calibration, and overall performance for a more detailed comparison. Values for censoring rate, sample sizes, and size of treatment effect are varied. 	\\

		\section{Materials and methods}
		
		\subsection{Reference datasets}
		Two clinical trial datasets serve as a reference for data simulations. The first dataset does not have any known treatment-covariate interactions (Section \ref{data_pbc}), and the second one comprises multiple  treatment-covariate interactions (Section \ref{data_pc}). 
		
		\subsubsection{Data without treatment-covariate interactions: Randomized Controlled Trial (RCT) in primary biliary cirrhosis}\label{data_pbc}
		An RCT conducted by the Mayo Clinic between 1974 and 1984 \citep{UMASS} investigates the effect of D-penicillamine on survival times in 312 patients with primary biliary cirrhosis (PBC), with time to the occurrence of death, or liver transplantation, respectively, as the event of interest. A total of 16 prognostic factors were recorded of which  ten were continuous and six were categorical variables. The median follow-up time is about five years. Table \ref{mayo_summary} shows more detailed summary statistics. We replaced missing values in three  of the continuous covariates by their column means, i.e. incomplete data are included for estimating the correlation structure and fitting univariate parametric distributions to the data.
		
		\begin{table}[htb]
			\small
			\renewcommand{\arraystretch}{1.25}
			
			\renewcommand\thetable{1a}
			
			\resizebox{\textwidth}{!}{ 
				\begin{tabular}{p{1em}lp{5em}p{8em}p{5em}p{5em}c} \toprule
					&		& {\small Median}            & {\small Mean (SE)}             & {\small Minimum}    & {\small Maximum}  & {\small \# missing values}  \\  \arrayrulecolor{black}\cmidrule[0.05pt]{1-7}
					\multicolumn{2}{l}{ {\small Survival time}} & & & & &  \\ \arrayrulecolor{gray!70}\cmidrule[0.05pt]{1-7}
					& Time of follow-up \textit{[Days]}  & 1839.5 & 2006.4 (1123.3) &  41 & 4556 & \textminus  \\ \arrayrulecolor{gray!70}\cmidrule[0.05pt]{1-7}
					\multicolumn{2}{l}{ {\small Continuous prognostic factors}} & & & & &  \\ \arrayrulecolor{gray!70}\cmidrule[0.05pt]{1-7}
					& Age \textit{[Years]}  &	 49.8 &  50 (10.6) & 26.3 & 78.4 & \textminus \\
					& Serum bilirubin \textit{[mg/dl]}  & 1.4 & 3.3 (4.5) & 0.3 & 28  & \textminus   \\   
					
					& Serum cholesterol  \textit{[mg/dl]}  & 322 & 369.6 (221.3) & 120 & 1775  & \textminus   \\ 
					
					& Albumin \textit{[gm/dl]} & 3.5 & 3.5 (0.4) & 2 & 4.6 &  \textminus  \\
					& Urine copper \textit{[mg/day]}	 & 73 & 97.6 (85.6) &  4 & 588 &  2 \\
					& Alkaline phosphatase \textit{[U/liter]}    & 1259 & 1982.7 (2140.4) & 289 & 13862.4 & \textminus \\ 
					& Aspartate aminotransferase  - SGOT \textit{[U/ml]} & 114.7 & 122.6 (56.7) & 26.4 & 457.2 &  \textminus \\
					& Triglycerides \textit{[mg/dl]}	  & 108 & 124.7 (65.1) & 33 & 598 & 30\\
					& Platelet count \textit{[\# platelets per m$^3$/1000]}	   & 257 & 261.9 (95.6) &  62 &  563 & 4 \\				
					& Prothrombin time \textit{[sec]}	   & 10.6 & 10.7 (1) &  9 & 17.1 & \textminus \\    \arrayrulecolor{black}\cmidrule[0.05pt]{1-7}
					&	& \multicolumn{4}{c}{ {\small Levels} }  &  {\small \# missing values}  \\  \arrayrulecolor{black}\cmidrule[0.05pt]{1-7}
					\multicolumn{2}{l}{ {\small Event indicator, treatment code}} & & & &  \\ \arrayrulecolor{gray!70}\cmidrule[0.05pt]{1-7}
					& Event indicator \textit{[0: censored, 1: death]}	& 0: 59.9\% & 1: 40.1\% & & & \textminus \\ 
					&  Treatment code \textit{[1: DPA, 2: placebo]}	& 1: 50.6\%  & 2: 49.4\% & & &  \textminus \\		\arrayrulecolor{gray!70}\cmidrule[0.05pt]{1-7}		
					\multicolumn{2}{l}{ {\small Categorical prognostic factors}} & & & &  \\ \arrayrulecolor{gray!70}\cmidrule[0.05pt]{1-7}
					& Sex \textit{[0: male, 1: female]}	& 0: 11.5\%  & 1: 88.5\% &&  &  \textminus \\
					& Presence of ascites \textit{[0: no, 1: yes]}	& 0: 92.3\%  & 1: 0.07\% & & &  \textminus \\
					& Presence of hepatomelagy \textit{[0: no, 1: yes]}	& 0: 48.7\%  & 1: 51.3\% & &  &  \textminus \\
					& Presence of spiders \textit{[0: no, 1: yes]}	& 0: 71.2\%  & 1: 28.8\% & & & \textminus \\
					& Presence of edema \textit{$^{1)}$}	& 0: 84.3\%  & 0.5: 9.3\% & 1: 6.4\% & & \textminus \\ 
					& Histologic state of disease \textit{[grade]}	& 1: 5.1\% & 2: 21.5\% & 3: 38.5\% & 4: 34.9\%  & \textminus \\  	\arrayrulecolor{black}\cmidrule[0.05pt]{1-7} 	
				\end{tabular}
			}	
			\caption[Summary statistics of baseline measurements in 312 primary biliary cirrhosis patients in the randomized controlled trial conducted by the Mayo Clinic.]{\small Randomized controlled trial in primary biliary cirrhosis: summary statistics of baseline measurements in 312 patients in the study conducted by the Mayo Clinic.
				\\{\footnotesize $^{1)}$ 0 = no edema and no diuretic therapy for edema; 0.5 = edema present for which no diuretic therapy was given or edema resolved with diuretic therapy; 1 = edema despite diuretic therapy.\\
					Abbreviations: DPA -  D-penicillamine, SGOT - serum glutamic-oxaloacetic transaminase }} 
			
			\label{mayo_summary}		
		\end{table}
		
		\noindent Performance comparison of the Cox model to a non-parametric alternative such as the RSF is motivated by the violation of the proportional hazard assumption in some datasets on which Cox regression is based. For instance, in this RCT dataset,  the overall assumption of proportional hazards would be violated ($\chi^2 = 20.86$, df = 8, $p = 0.0075$, test by Grambsch and Therneau \citep{Grambsch1994} implemented in the  function \texttt{cox.zph} from the \texttt{R}  package \texttt{survival} \citep{Therneau2024}) after variable selection based on   findings in the literature and the statistical measures AIC (Akaike information criterion) and BIC (Bayesian information criterion), an approach a researcher examining these data would typically follow.

		\subsubsection{Data with treatment-covariate interactions: Randomized Controlled Trial (RCT) in  prostate cancer patients}\label{data_pc}
		The second dataset considered comprises 474 patients with advanced prostate cancer for whom complete data are available in the RCT examining the effect of the synthetic oestrogen drug diethyl stilboestrol on survival time. The placebo group comprises patients receiving either placebo or the lowest dose level, the treatment group comprises patients receiving one of  two higher dose levels \citep{Byar1980}. Table \ref{byar_summary} gives an overview of the data structure.
		For data simulations, we removed the binary variable cancer stage due to multicollinearity. Based on findings in the literature  \citep{Byar1980, Royston2004}, we included relevant interaction terms between treatment and the variables age, presence of bone metastases, and serum acid phosphatase, respectively. Again, in a model comprising all main effects and these three interaction terms, for example, the proportional hazard assumption would not be fulfilled ($\chi^2 = 22.2$, df = 12, $p = 0.0355$,  test by Grambsch and Therneau \citep{Grambsch1994}).

		\begin{table}[htb]
			\small
			\renewcommand{\arraystretch}{1.25}
			\renewcommand\thetable{1b}
			\resizebox{\textwidth}{!}{ 
				\begin{tabular}{p{0.5em}lp{5em}p{8em}p{5em}p{5em}c} \toprule
					&		& {\small Median}            & {\small Mean (SE)}             & {\small Minimum}    & {\small Maximum}  & {\small \# missing values}  \\  \arrayrulecolor{black}\cmidrule[0.05pt]{1-7}
					\multicolumn{2}{l}{ {\small Survival time}} & & & & &  \\ \arrayrulecolor{gray!70}\cmidrule[0.05pt]{1-7}
					& Time of follow-up \textit{}  & 33.5 & 36.3 (23.2) &  0.5 &  76.5 & \textminus  \\ \arrayrulecolor{gray!70}\cmidrule[0.05pt]{1-7}
					\multicolumn{2}{l}{ {\small Continuous prognostic factors}} & & & & &  \\ \arrayrulecolor{gray!70}\cmidrule[0.05pt]{1-7}
					& Age \textit{[Years]}         &	 73 &  71.6 (6.9) & 48 & 89 & \textminus \\
					& Standardized weight \textit{}  & 98 & 99 (13.3) & 69 & 152  & \textminus   \\   
					
					& Systolic blood pressure  \textit{}       & 14 & 14.4 (2.4) & 8 & 30  & \textminus   \\ 
					
					& Diastolic blood pressure \textit{}        & 8 & 8.2 (1.5) & 4 & 18 &  \textminus  \\
					& Size of primary tumour \textit{[cm$^2$]}	 & 10 & 14.3 (12.2) &  0 & 69 & \textminus \\
					& \parbox{6cm}{Serum (prostatic) acid phosphatase \\ \textit{[King Armstrong units]}}   & 7 & 125.7 (638.5) & 1 & 9999  & \textminus \\ 
					& Haemoglobin \textit{[g/100 ml]}                                      & 137 & 134.2 (19.4) & 59 & 182 &  \textminus \\
					& Gleason stage-grade category \textit{[mg/dl]}	                       & 10 & 10.3 (2) & 5 & 15 & \textminus\\   \arrayrulecolor{black}\cmidrule[0.05pt]{1-7}
					&	& \multicolumn{4}{c}{ {\small Levels} }  &  {\small \# missing values}  \\  \arrayrulecolor{black}\cmidrule[0.05pt]{1-7}
					\multicolumn{2}{l}{ {\small Event indicator, treatment code}} & & & &  \\ \arrayrulecolor{gray!70}\cmidrule[0.05pt]{1-7}
					& Event indicator \textit{[0: censored, 1: death]}        	& 0: 28.8\% & 1: 71.2\% & & & \textminus \\ 
					&  	\parbox{6cm}{Treatment code \\ \textit{[0: lowest dose of diethyl stilboestrol (placebo), 1: higher doses]}}	            & 0: 49.9\%  & 1: 50.1\% & & &  \textminus \\		\arrayrulecolor{gray!70}\cmidrule[0.05pt]{1-7}		
					\multicolumn{2}{l}{ {\small Binary prognostic factors}} & & & &  \\ \arrayrulecolor{gray!70}\cmidrule[0.05pt]{1-7}
					& Performance status \textit{}	                    & 0: 90.1\%  & 1: 9.9\% &&  &  \textminus \\
					& History of cardiovascular disease  \textit{[0: no, 1: yes]}	& 0: 56.6\%  & 1: 43.4\% & & &  \textminus \\
					& Presence of bone metastases \textit{[0: no, 1: yes]}	        & 0: 83.8\%  & 1: 16.2\% & &  &  \textminus \\
					& \parbox{6cm}{Abnormal electrocardiogram \\ \textit{[0: normal, 1: abnormal]}}	& 0: 34.1\%  & 1: 65.9\% & & & \textminus \\
					\arrayrulecolor{black}\cmidrule[0.05pt]{1-7} 	
				\end{tabular}
			}
			\caption[Summary statistics of baseline measurements in 474 prostate cancer patients in the randomized controlled trial dataset in prostate cancer patients.]{\small{Randomized controlled trial in prostate cancer patients: summary statistics of baseline measurements in 474 patients in the prostate cancer dataset.} }
			\label{byar_summary}
		\end{table}

		\subsection{Methods for performance comparison}
		The approaches were first compared using the bootstrap technique by Wahl et al. (\citeyear{Wahl2016}) which is based on the work by Jiang et al. (\citeyear{Jiang2008}). This is an internal validation technique based on the real data for estimating point estimates of the performance measures and corresponding CIs. It is described in Section \ref{method_boot}. 
		Moreover, we used data simulations which facilitate manipulations of data properties but  at the same time require specification of data-generating mechanisms. 
		The approach is described in Section \ref{method_sim}. Model building for finding the most suitable model for each dataset, was done in the same way, for both the bootstrap and simulated data. Details are given in Supplementary Material A.
		
		\subsubsection{Nonparametric bootstrap approach}\label{method_boot}

		\noindent	The nonparametric bootstrap approach for point estimates by Wahl et al. (\citeyear{Wahl2016}) is an extension of the algorithm by Jiang et al. (\citeyear{Jiang2008}) and based on the .632+ bootstrap method \citep{Efron1997}.
		\\
		The .632+ bootstrap estimate ($\hat{\theta}^{.632+}$) of the performance measure of interest  is computed as a weighted average of the apparent performance $\hat{\theta}^{orig,orig}$ (training and test data given by the original dataset) and the average ``out-of-bag'' (OOB) performance $\hat{\theta}^{bootstrap,OOB} = \sum\limits_{b=1}^B \hat{\theta}^{bootstrap,OOB}_b$ computed from $B$ bootstrap datasets (training data given by the bootstrap dataset, and test data given by the samples not present in the bootstrap dataset). The formula is:
		\begin{equation*}
			\hat{\theta}^{.632+} = (1-w) \cdot \hat{\theta}^{orig,orig} + w \cdot \hat{\theta}^{bootstrap,OOB},
		\end{equation*}
		where $w = \frac{0.632}{1-0.368 \cdot \text{R}}$ and R = $\frac{\hat{\theta}^{bootstrap,OOB} - \hat{\theta}^{orig,orig}}{\theta^{noinfo}   - \hat{\theta}^{orig,orig}}$. In case of the $C$ index, $\theta^{noinfo} = 0.5$. For the Integrated Brier score, $\theta^{noinfo} = 0.75$. 
		Then each bootstrap dataset is assigned a weight $w_b =  \hat{\theta}^{bootstrap,bootstrap}_b - \hat{\theta}^{orig,orig}$, where $\hat{\theta}^{bootstrap,bootstrap}_b$ is the value of the performance measure, when the bootstrap dataset $b \in \{1,\cdots,B\}$ is used as training as well as test dataset. The  $\frac{\alpha}{2}$ and $1 - \frac{\alpha}{2}$ percentiles of the empirical distribution of these weights, $\xi_{\frac{\alpha}{2}}$ and $\xi_{1 - \frac{\alpha}{2}}$ , give the CI of $\hat{\theta}^{.632+}$:
		\begin{equation*}
			[\hat{\theta}^{.632+} -  \xi_{1 - \frac{\alpha}{2}}, \hat{\theta}^{.632+} + \xi_{\frac{\alpha}{2}}]
		\end{equation*}

		\subsubsection{Data simulation}\label{method_sim}
		For data simulations, covariate data similar to the reference data are generated by using a copula model. The specific  distributions and corresponding parameters can be found in the Supplementary Material B: 
		Table B.1a 
		and Table B.1b 
		show the correlation matrices estimated from the primary biliary cirrhosis and the prostate cancer dataset, respectively.  
		Table B.2a 
		and Table B.2b 
		show the assumed parametric distributions of each variable in both reference datasets.  In Figure B.1 
		and Figure B.2 
		the empirical variable distributions and the best fitting theoretical distributions based on maximum likelihood estimation are shown for the primary biliary cirrhosis and the prostate cancer dataset, respectively.
		Covariate-dependent survival times were generated from a Cox proportional hazards model assuming Weibull($\lambda, \gamma$) distributed survival times according to the cumulative hazard inversion method by Bender et al. (\citeyear{Bender2005})  implemented in the \texttt{R} package \texttt{simsurv} \citep{Brilleman2022}. 
		For this,  regression parameters $\bm{\beta}$ were estimated based on the reference datasets. For the PBC dataset
		
		\begin{equation*}
			\begin{aligned}
				\bm{\beta}^{\text{PBC}} = & (\beta_1, \dots, \beta_{17}) \\		
				= & (\beta_{Z1}, \beta_{Z2}, \beta_{Z3}, \beta_{Z4}, \beta_{Z5}, \beta_{Z6}, \beta_{Z7}, \beta_{Z8}, \\
				&  \beta_{Z9}, \beta_{Z10}, \beta_{Z11}, \beta_{Z12}, \beta_{Z13}, \beta_{Z14}, \beta_{Z15}, \beta_{Z16}, \beta_{Z17})    \\
				\approx &   (\beta_{Z1},\,  0.026,\, -0.218,\,  0.338,\,  0.227, \, 0.071,\,  0.481,\,  0.086, \\
				&  0.0004, \, -0.799, \, 0.003,\,  -0.00002,\,  0.004,\,  -0.002,\,  0.0002,\,  0.276,\, 0.365).
			\end{aligned}
		\end{equation*}

		For the prostate cancer dataset
		\begin{equation*}
			\begin{aligned}
				\bm{\beta}^{\text{PC}} = & (\beta_1, \dots, \beta_{16}) \\		
				= & (\beta_{RX}, \beta_{AGE}, \beta_{WT}, \beta_{SBP}, \beta_{DBP}, \beta_{SZ}, \beta_{AP}, \beta_{HG}, \\
				& \beta_{SG}, \beta_{PF}, \beta_{HX}, \beta_{BM}, \beta_{ECG}, \beta_{RX:AGE}, \beta_{RX:BM}, \beta_{RX:AP} )    \\
				\approx &   (\beta_{RX},\, -0.006,\, -0.01,\, -0.016,\,  0.02,\,  0.014, \, 0.0001,\, -0.006,\\
				&   0.074, \, 0.333, \, 0.467, \, 0.63, \, 0.316, \, 0.059,\, -0.612, \, -0.0003).
			\end{aligned}
		\end{equation*}

		\noindent Scale parameters $\lambda$ were fixed at the value estimated from the respective reference dataset ($\lambda$ = 2241.74 for the primary biliary cirrhosis dataset, $\lambda$ = 39.2 for the prostate cancer dataset), shape parameters $\gamma$ were varied in order to create scenarios with decreasing ($\gamma = 0.8$), constant ($\gamma = 1$), increasing ($\gamma = 2$), and non-proportional hazards, i.e. different values per treatment group ($\gamma_0 = 2, \gamma_1 = 5$).  Random censoring times were generated from a uniform distribution $U_{[0,b]}$ such that censoring percentages of 30\% and 60\%, respectively, corresponding to the actual censoring rates in the two reference datasets, were obtained. For this, the approach by Ramos et al. (\citeyear{Ramos2024}) was used, but in some cases  the values of the distribution parameter $b$ had to be manually adjusted. Total sample sizes $N \in \{100, 200, 400\}$ were considered for the $n_{\text{sim}} = 500$ training datasets. For the $n_{\text{sim}} = 500$ independent test datasets, the total sample size is $N = 500$. In contrast to analysing real-world data, where the available observations are split into a training and test dataset, possibly several times in order to perform cross-validation depending on the size of the dataset, simulations do not rely on actual data. Thus, independent test datasets can be generated \citep{Graf2025}. In simulations based on time-to-event data, this additionally provides the advantage of maintaining a certain censoring rate in both, the training and test data. Moreover, different values of the treatment effect ($\beta_{\text{treatment}} \in \{0, 0.8,  -0.4\}$) were considered when generating the data  corresponding to different hazard ratios of the treatment effect.
		For the RSF, all available splitting rules  are included in the method comparison (overview in Table A.1.   
		). 
		Computational times per algorithm were measured including variable selection (for the Cox model) and hyperparameter tuning (for the RSF model), respectively.

		\subsection{Performance measures}\label{perf_meas}
		According to recommendations, performance metrics measuring discrimination, calibration, and overall performance  shall be reported when comparing prediction models. In the context of survival analysis, discrimination refers to the model's ability to distinguish between patients with higher and lower risk of the outcome. Calibration compares predicted survival probabilities to the observed event frequencies in a given time interval. Overall performance encompasses both discrimination as well as calibration of the model. Some performance measures have been extended for use with survival outcomes.

		\subsubsection{Measure of discrimination: Harrell's $C$ index}\label{cindex}
		The $C$ index was originally developed for binary outcomes \citep{Harrell1985}, and has been subject to criticism \citep{Hartman2023}. It compares for each pair of patients  whether the one with the shorter event time also has the higher predicted risk of suffering the event.  These rank-based comparisons may favour the model with the more inaccurate predictions \citep{Vickers2010}, and may not adequately reflect the influence  different sets of covariates have on the outcome \citep{Cook2007}, such that its interpretation may be misleading and not clinically meaningful for survival outcomes.\\
		The $C$ index, a time range measure, can be obtained from the Cox regression and RSF models as follows. For a Cox proportional hazards model
		\begin{equation*}
			h(t) = h_0(t)\exp(\beta_1 x_1 + \dots + \beta_{d} x_d)  
		\end{equation*}
		with baseline hazard function $ h_0(t)$ and regression coefficients $\bm{\beta} \in \mathds{R}^d$, and unique ordered survival times $t_1,\dots,t_m$, at each uncensored survival time, the rank of the predicted outcome for the considered subject $j_1$ who experienced the event, i.e. $\hat{h}_{j_1}(t)$, is compared to all $\hat{h}_{j_2}(t), j_1 \neq j_2$ where individuals $j_2$ had a longer survival time. The $C$ index can thus be written as:
		\begin{equation*}
			\text{Pr}(\hat{h}_{j_1} > \hat{h}_{j_2}|T_{j_1} < T_{j_2}) = \frac{\sum\limits_{j_1} (R_{j_1} - 1)}{\sum\limits_{j_1} (N_{j_1} - 1)}, \hspace{0.5cm} j_1,j_2 \in \{1,\dots m\}, j_1 \neq j_2
		\end{equation*} 
		where $R_{j_1}$ is the rank of individual $j_1$ with survival time $T_{j_1}$, $N_{j_1}$ the number at risk at time $T_{j_1}$, and thus $N_{j_1} -1$ the number of individuals who can be compared with $j_1$   \citep{Kremers2007}. \\
		For the RSF model, the $C$ index is computed based on the patient-specific predictions of the ensemble mortality in the terminal nodes of each tree. For this the out-of-bag (oob) ensemble estimator of the cumulative hazard function (CHF) at time $t$ for patient $j$, $H_j^{oob} (t)$, is considered. It is given by the average prediction of the $n_{\text{tree}_j}$ trees for which the sample was not part of the bootstrap sample for building the tree \citep{Ishwaran2021a}:
		\begin{equation*}
			H_j^{oob} (t) = \frac{1}{n_{\text{tree}_j}} \sum_{b \in n_{\text{tree}_j}} H_b(t|\mathbf{X})
		\end{equation*}  
		where $H_b(t|\mathbf{X})$ is the CHF predicted in the terminal node of the $b$th tree for the covariate vector $\mathbf{X} \in \mathds{R}^d$ of patient $j$ at time $t$.
		The out-of-bag ensemble mortality for each patient $j = 1,\dots,N$ is then estimated as the sum of the oob CHF estimates over all unique event times $t_1,\dots,t_m$ in the training data \citep{Ishwaran2021a}:
		\begin{equation*}
			M_j^{oob} = \sum_{k=1}^m H_j^{oob}(t_k)
		\end{equation*}
		The $C$ index is the proportion of concordant pairs among all pairs for which the decision can be made. If $M_{j_1}^{oob} >  M_{j_2}^{oob}$ and patient $j_1$ has the shorter event time compared to patient $j_2$ or vice versa, the pair is concordant. The closer $C$ index estimates are to 1 the better.

		\subsubsection{Measure of calibration: Calibration curves}
		A calibration plot of observed on predicted probabilities of mortality indicates deviation from perfect prediction the more the slope deviates from the ideal  line with slope 1 \citep{Calster2019}. It quantifies
		the agreement between the actual and predicted outcome within a specified duration of time. Austin et al. (\citeyear{Austin2020}) describe and implement an approach for estimating calibration curves for survival outcomes. The calibration curve used here is estimated based on Cox regression using restricted cubic splines.

		\subsubsection{Measure of overall performance: Integrated Brier score}	
		The Integrated Brier score \citep{Graf1999} summarizes the Brier scores over time, i.e. it is a time range performance measure.The Brier score calculates the difference between predicted and actual survival at a given time point, and thus values indicate better overall performance the closer they are to zero. It is implemented in the function \texttt{integrated$\_$brier$\_$score} in the \texttt{R} package \texttt{survex} \citep{survex2024}.

		\section{Results}
		
		\subsection{Results of the bootstrap approach}

\begin{figure}[H]
			\begin{minipage}{.47\textwidth}
				\includegraphics[scale = 0.5]{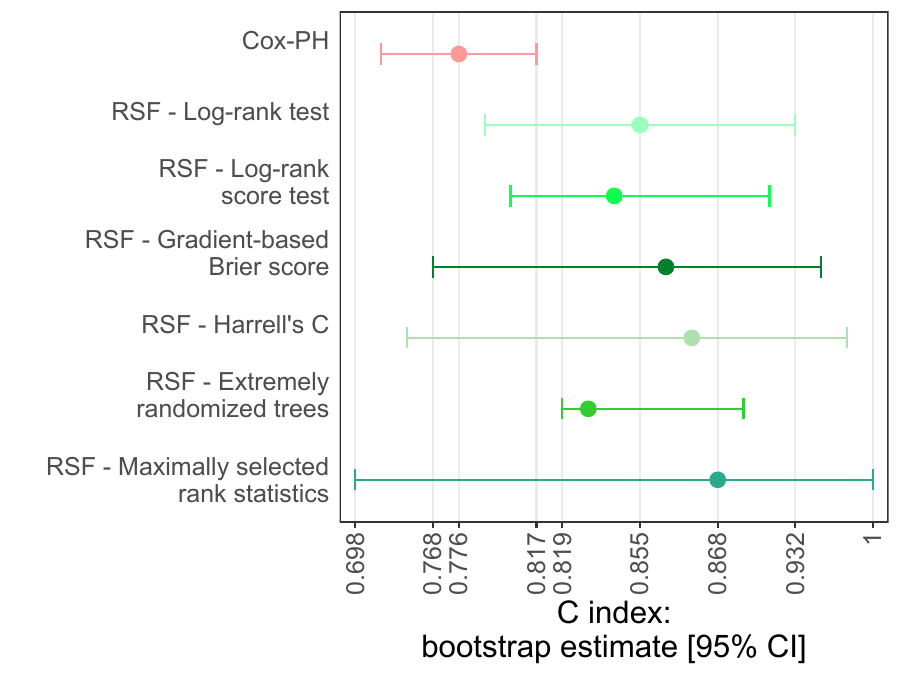}
			\end{minipage}
			\begin{minipage}{.05\textwidth}
				\qquad 			
			\end{minipage}
			\begin{minipage}{.47\textwidth}
				\includegraphics[scale = 0.5]{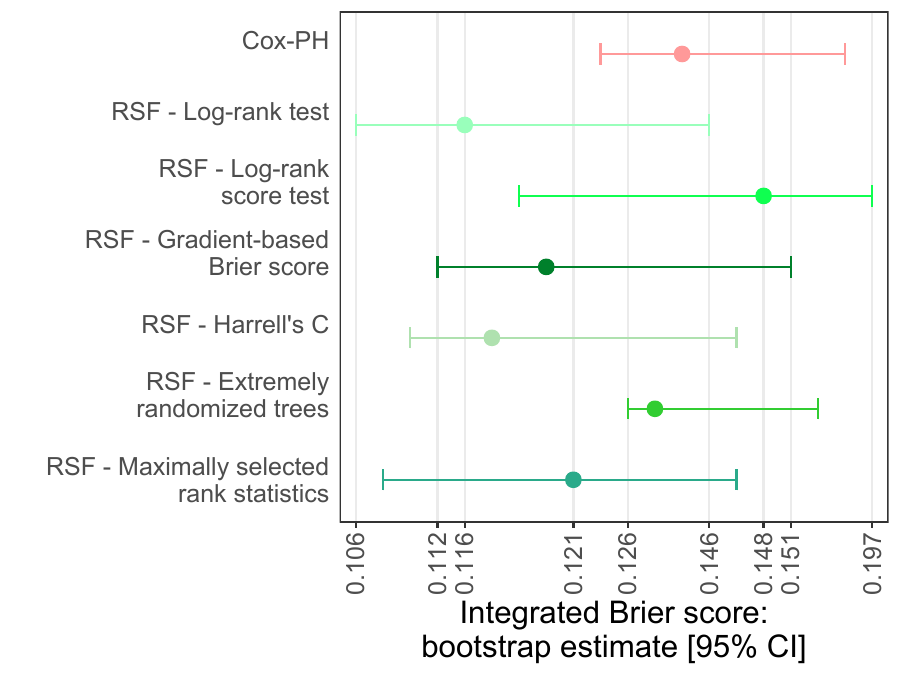}					
			\end{minipage}
			
			\vspace{0.1cm}
			\captionof{figure}[Bootstrap estimate $\hat{\theta}^{.632+}$ (95\% confidence interval) of the $C$ index and Integrated Brier score for the RCT in primary biliary cirrhosis patients.]{\fontsize{9}{10}\selectfont \small{\textbf{Bootstrap estimate $\hat{\theta}^{.632+}$ (95\% confidence interval) of the $C$ index (right) and Integrated Brier score (left) for the RCT in  \underline{primary biliary cirrhosis} patients. \\  } } 			
				{  \footnotesize  Abbreviations: Cox-PH - Cox proportional hazards model, RSF - Random survival forest.}		
			}
			
			\label{mayo_boot_plot}
			
		\end{figure}

\begin{figure}[H]
			\begin{minipage}{.47\textwidth}
				\includegraphics[scale = 0.5]{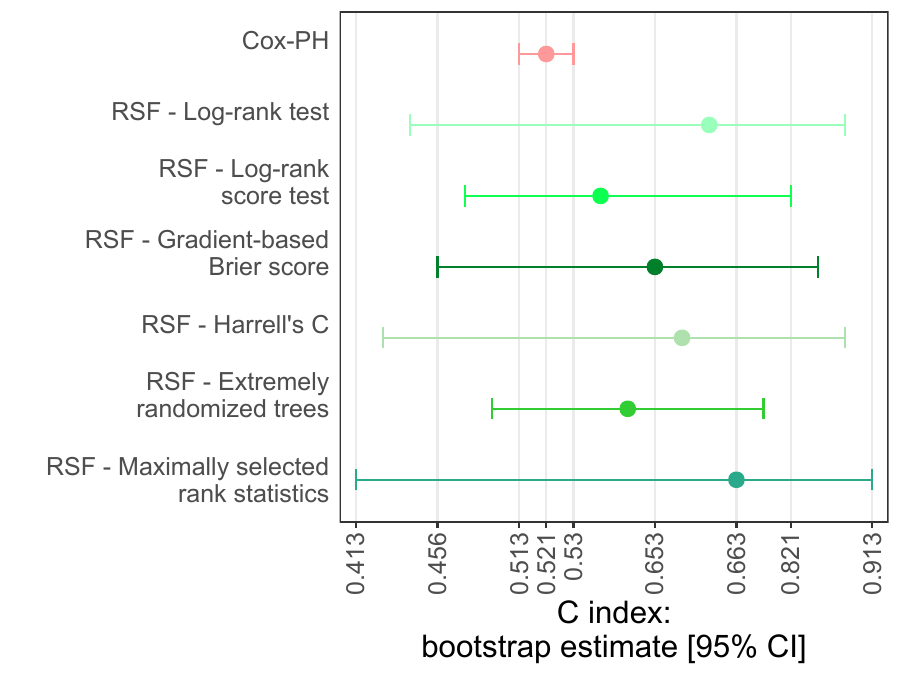}
			\end{minipage}
			\begin{minipage}{.05\textwidth}
				\qquad 			
			\end{minipage}
			\begin{minipage}{.47\textwidth}
				\includegraphics[scale = 0.5]{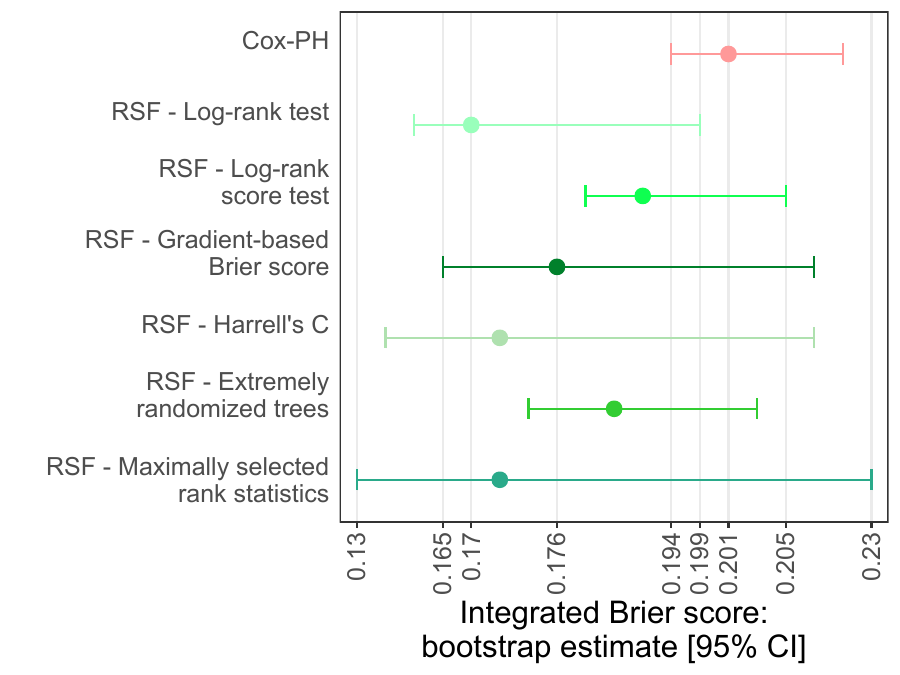}					
			\end{minipage}
			
			\vspace{0.1cm}
			\captionof{figure}[Bootstrap estimate $\hat{\theta}^{.632+}$ (95\% confidence interval) of the $C$ index and Integrated Brier score for the RCT data in primary biliary cirrhosis patients.]{\fontsize{9}{10}\selectfont \small{\textbf{Bootstrap estimate $\hat{\theta}^{.632+}$ (95\% confidence interval) of the $C$ index (left) and Integrated Brier score (right) for the RCT in \underline{prostate cancer} patients.\\}}{\footnotesize  Abbreviations: Cox-PH - Cox proportional hazards model, RSF - Random survival forest.}		
			}
			
			\label{byar_boot_plot}		
		\end{figure}

		Table \ref{bootstrap_mayo} and Table \ref{bootstrap_byar} show the bootstrap estimates of the $C$ index and Integrated Brier score, respectively, when applying the bootstrap approach for point estimates \citep{Jiang2008, Wahl2016} to both reference datasets. The same results are shown in Figure \ref{mayo_boot_plot} (primary biliary cirrhosis dataset) and Figure \ref{byar_boot_plot} (prostate cancer dataset).
		The first impression is that the point estimates $\hat{\theta}^{.632+}$ alone indicate a potentially better performance of most RSF models compared to the Cox-PH model but their confidence intervals are often much wider and mostly include the $\hat{\theta}^{.632+}$ estimate of the Cox-PH model. This is the case for both reference datasets. \\ 
		The RSF (``log-rank test'') may have an advantage over the Cox model when comparing overall performance (using the Integrated Brier score) in both datasets, because its point estimates $\hat{\theta}^{.632+}$ are the lowest and the corresponding confidence intervals have the smallest overlap with the confidence interval belonging to $\hat{\theta}^{.632+}$ of the Cox-PH model. With respect to the $C$ index, the RSF (``extremely randomized trees'') performs better in the data without treatment-covariate interactions (primary biliary cirrhosis dataset).  Confidence intervals are non-overlapping for this apporach and the Cox-PH model. For the prostate cancer dataset, RSF may have better performance concluded from the point estimates alone, but confidence intervals of the Cox and RSF models completely overlap such that no clear conclusion can be made.

		
		\begin{table}[htb]
			\renewcommand\thetable{2}
			\def\arraystretch{0.6}
			\setlength{\tabcolsep}{6pt}
			\caption[Bootstrap estimates $\hat{\theta}^{.632+}$ (95\% confidence interval) of the $C$ index and Integrated Brier score in the  data without treatment-covariate interactions (primary biliary cirrhosis dataset).]{\fontsize{9}{10}\selectfont  Bootstrap estimates $\hat{\theta}^{.632+}$ (95\% confidence interval) of the $C$ index and Integrated Brier score in the RCT data without treatment-covariate interactions (\underline{primary biliary cirrhosis dataset}). Predictions are based on $n_{\text{sim}}$ = 1000 bootstrap datasets.}		
			\label{bootstrap_mayo}		
			\footnotesize		
			\begin{tabular}{@{}p{4em}p{6em}p{6em}p{6em}p{6em}p{6em}p{6em}p{6em}@{}} \arrayrulecolor{black}\cmidrule[0.1pt]{1-8} 
				
				& \multirow{4}{*}{Cox-PH}  & \multicolumn{6}{c}{\small{Random survival forest}} \\ \arrayrulecolor{black}\cmidrule[0.05pt]{3-8}
				&                     & \parbox{4em}{\linespread{1}\selectfont Log-rank test} & \parbox{4em}{\linespread{1}\selectfont Log-rank score} & \parbox{7em}{\linespread{1}\selectfont Gradient-based Brier score} & \parbox{5em}{\linespread{1}\selectfont Harrell's $C$} & \parbox{6em}{\linespread{1}\selectfont Extremely randomized trees} & \parbox{6em}{\linespread{1}\selectfont Maximally selected rank statistics} \\ \arrayrulecolor{black}\cmidrule[0.05pt]{1-8} 
				
				$C$ index                                                        & 0.776 (0.735,0.817) & 0.855 (0.778,0.932)                                   & 0.847 (0.815,0.88)                                     & 0.856 (0.768,0.944)                                                & 0.858 (0.764,0.953)                                   & 0.844 (0.819,0.869)                                                & 0.868 (0.698,1)                                                            
				\\ \arrayrulecolor{gray!50}\cmidrule[0.01pt]{1-8}
				\parbox{4em}{\linespread{1}\selectfont  Integrated Brier  score} & 0.131 (0.124,0.161) & 0.116 (0.106,0.146)                                   & 0.148 (0.118,0.197)                                    & 0.12 (0.112,0.151)                                                 & 0.117 (0.109,0.147)                                   & 0.129 (0.126,0.152)                                                & 0.121 (0.108,0.147)                                                        
				\\   \arrayrulecolor{black}\arrayrulecolor{black}\cmidrule[0.1pt]{1-8}		 
			\end{tabular}
		\end{table}
		
		\vspace{-1cm}
		
		\vspace{-1cm}
		
		\begin{table}[htb]
			\renewcommand\thetable{3}
			\def\arraystretch{0.6}
			\setlength{\tabcolsep}{6pt}
			\caption[Bootstrap estimates $\hat{\theta}^{.632+}$ (95\% confidence interval) of the $C$ index and Integrated Brier score in the data with three treatment-covariate interactions (prostate cancer dataset).]{\fontsize{9}{10}\selectfont  Bootstrap estimates $\hat{\theta}^{.632+}$ (95\% confidence interval) of the $C$ index and Integrated Brier score  data with three treatment-covariate interactions (\underline{prostate cancer dataset}). Predictions are based on $n_{\text{sim}}$ = 1000 bootstrap datasets.}
			\label{bootstrap_byar}		
			\footnotesize		
			\begin{tabular}{@{}p{4em}p{6em}p{6em}p{6em}p{6em}p{6em}p{6em}p{6em}@{}} \arrayrulecolor{black}\cmidrule[0.1pt]{1-8} 
				
				& \multirow{4}{*}{Cox-PH}  & \multicolumn{6}{c}{\small{Random survival forest}} \\ \arrayrulecolor{black}\cmidrule[0.05pt]{3-8}
				&                     & \parbox{4em}{\linespread{1}\selectfont Log-rank test} & \parbox{4em}{\linespread{1}\selectfont Log-rank score} & \parbox{7em}{\linespread{1}\selectfont Gradient-based Brier score} & \parbox{5em}{\linespread{1}\selectfont Harrell's $C$} & \parbox{6em}{\linespread{1}\selectfont Extremely randomized trees} & \parbox{6em}{\linespread{1}\selectfont Maximally selected rank statistics} \\ \arrayrulecolor{black}\cmidrule[0.05pt]{1-8} 
				
				$C$ index                                                        & 0.521 (0.513,0.53)  & 0.66 (0.438,0.881)                                    & 0.642 (0.462,0.821)                                    & 0.653 (0.456,0.85)                                                 & 0.657 (0.432,0.881)                                   & 0.645 (0.498,0.792)                                                & 0.663 (0.413,0.913)                                                        \\ \arrayrulecolor{gray!50}\cmidrule[0.01pt]{1-8}
				\parbox{4em}{\linespread{1}\selectfont  Integrated Brier  score} & 0.201 (0.194,0.211) & 0.17 (0.158,0.199)                                    & 0.183 (0.177,0.205)                                    & 0.176 (0.165,0.206)                                                & 0.172 (0.155,0.206)                                   & 0.179 (0.175,0.204)                                                & 0.172 (0.13,0.23)                                                          
				\\   \arrayrulecolor{black}\arrayrulecolor{black}\cmidrule[0.1pt]{1-8}		 
			\end{tabular}
		\end{table}
		
		\vspace{-1cm}
		
		\subsection{Simulation study results}
				
\begin{figure}[H]
					\begin{subfigure}{\textwidth}
						\begin{minipage}{0.98\textwidth} 	
							\renewcommand\thefigure{}
							\renewcommand{\figurename}{}
							\includegraphics[scale = 0.2, width = 0.995\textwidth,trim={0cm, 0, 0cm, 0},clip]{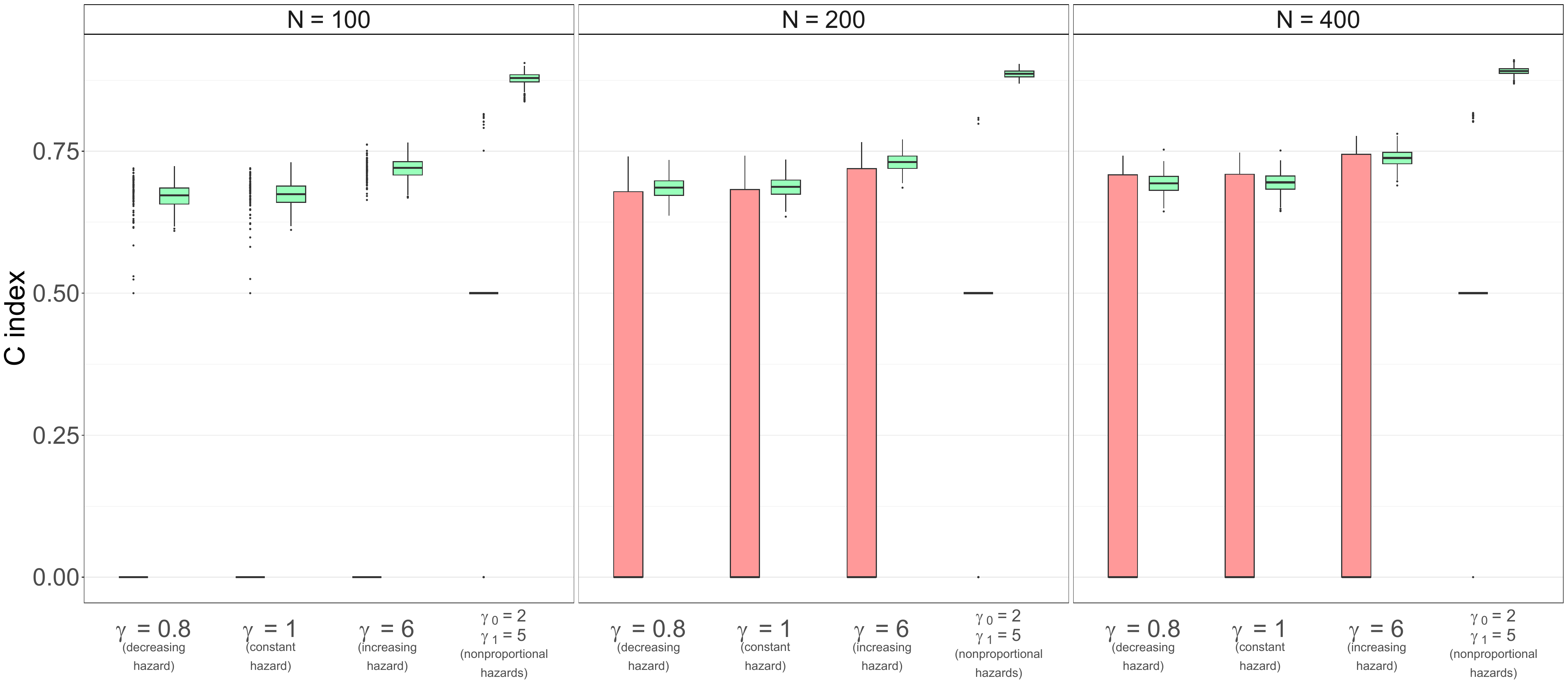}   
							\vspace{-0.7cm}
							\caption[]{\textbf{}}		
						\end{minipage}		
						\begin{minipage}{0.005\textwidth} 	
							\includegraphics[scale = 0.325,trim={0cm, 0, 0cm, 0},clip]{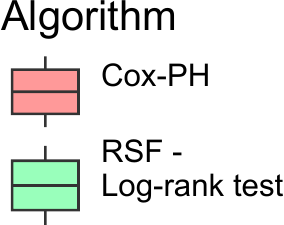}
						\end{minipage}	
					\end{subfigure}		
					\begin{subfigure}{\textwidth}	
						\begin{minipage}{0.98\textwidth} 	
							\renewcommand\thefigure{}
							\renewcommand{\figurename}{}
							\includegraphics[scale = 0.2, width = 0.995\textwidth,trim={0cm, 0, 0cm, 0},clip]{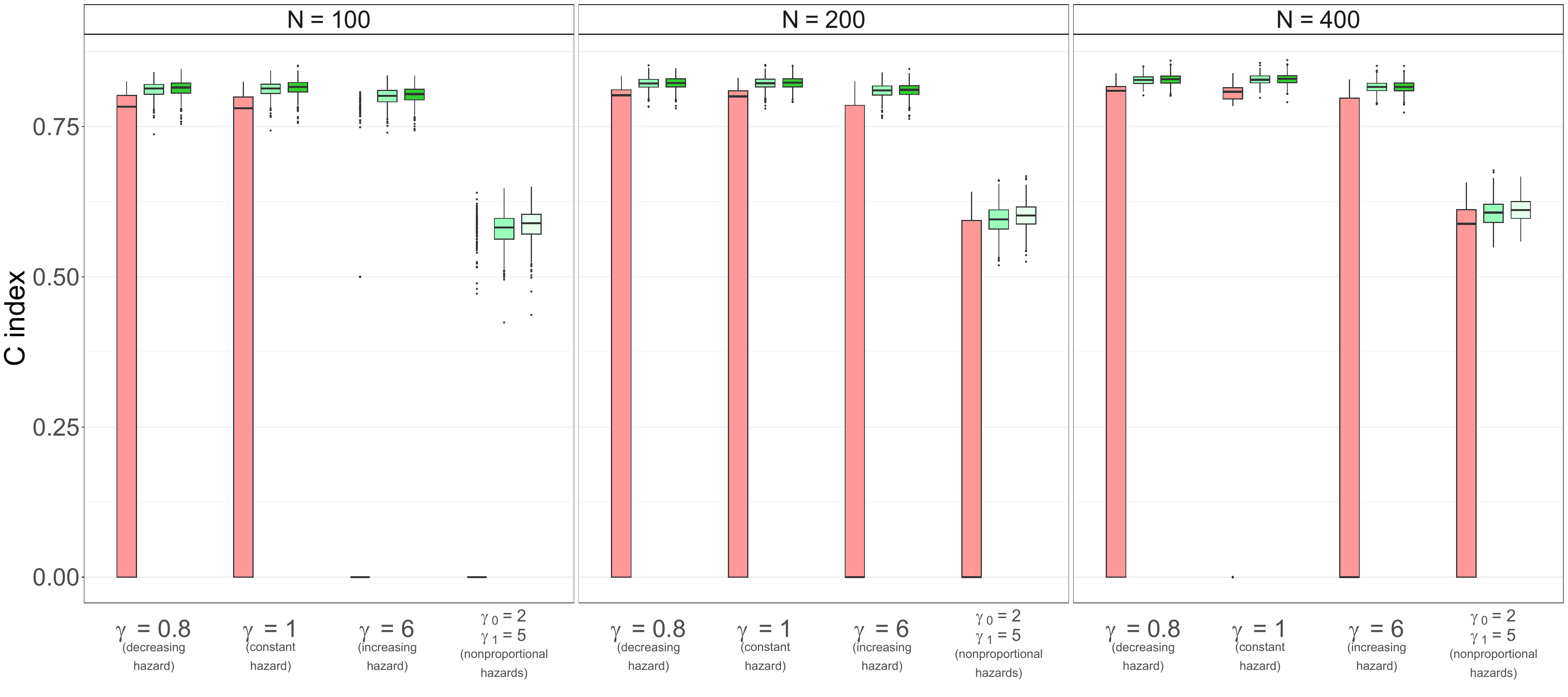}   
							\vspace{-0.7cm}
							\caption[]{\textbf{}}
						\end{minipage}		
						\begin{minipage}{0.005\textwidth} 	
							\includegraphics[scale = 0.325,trim={0cm, 0, 0cm, 0},clip]{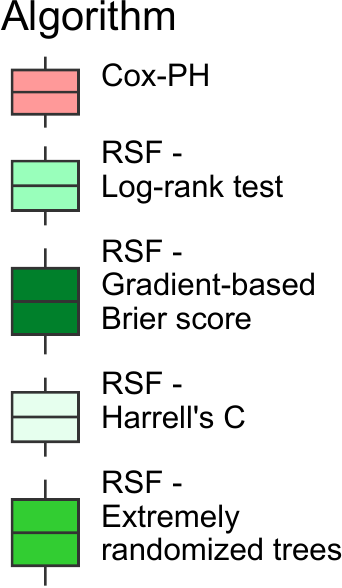}
						\end{minipage}	
					\end{subfigure}	
					\vspace{-0.3cm}
					\caption[$C$ index estimates in the simulation scenario assuming 30\% censoring and a treatment effect of $\beta_{\text{\normalfont{treatment}}} = -0.4$ for the RCT in primary biliary cirrhosis (a) and in prostate cancer patients (b).]{
						{\small \textbf{$C$ index estimates in the simulation scenario assuming 30\% censoring and a treatment effect of $\beta_{\text{\normalfont{treatment}}} = -0.4$ for the RCT in primary biliary cirrhosis (a) and in prostate cancer patients (b).} Survival times are generated from a Weibull distribution with scale parameters estimated from the respective reference dataset, shape parameters ($\gamma$) vary in order to examine the impact of differing hazards, and the violation of the proportional hazards assumption.
							Results are shown for different total sample sizes $N$.\\}
						{\footnotesize  Abbreviations:  Cox-PH - Cox proportional hazards model, RSF - Random survival forest.}
					} 
					\label{boxplots_cindex_beta0_cens_30}				
				\end{figure}

\begin{figure}[H]

					\begin{subfigure}{\textwidth}
						\begin{minipage}{0.98\textwidth} 	
							\renewcommand\thefigure{}
							\renewcommand{\figurename}{}
							\includegraphics[scale = 0.2, width = 0.995\textwidth,trim={0cm, 0, 0cm, 0},clip]{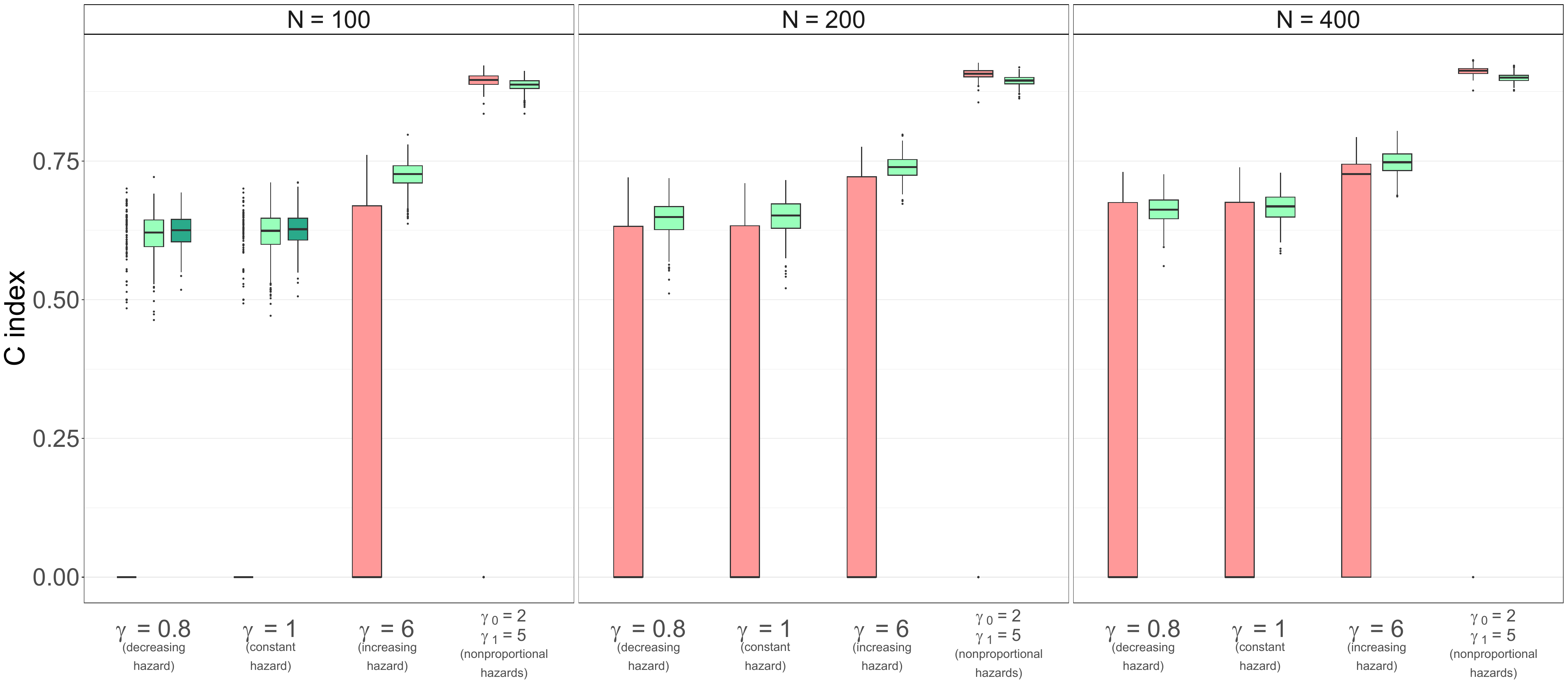}   
							\vspace{-0.7cm}
							\caption[]{\textbf{}}		
						\end{minipage}		
						\begin{minipage}{0.005\textwidth} 	
							\includegraphics[scale = 0.325,trim={0cm, 0, 0cm, 0},clip]{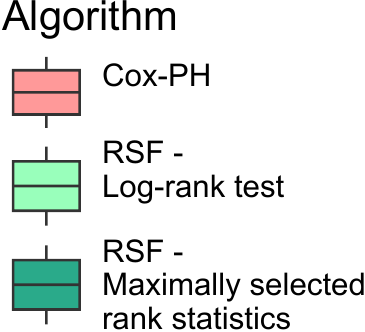}
						\end{minipage}	
					\end{subfigure}		
					\begin{subfigure}{\textwidth}	
						\begin{minipage}{0.98\textwidth} 	
							\renewcommand\thefigure{}
							\renewcommand{\figurename}{}
							\includegraphics[scale = 0.2, width = 0.995\textwidth,trim={0cm, 0, 0cm, 0},clip]{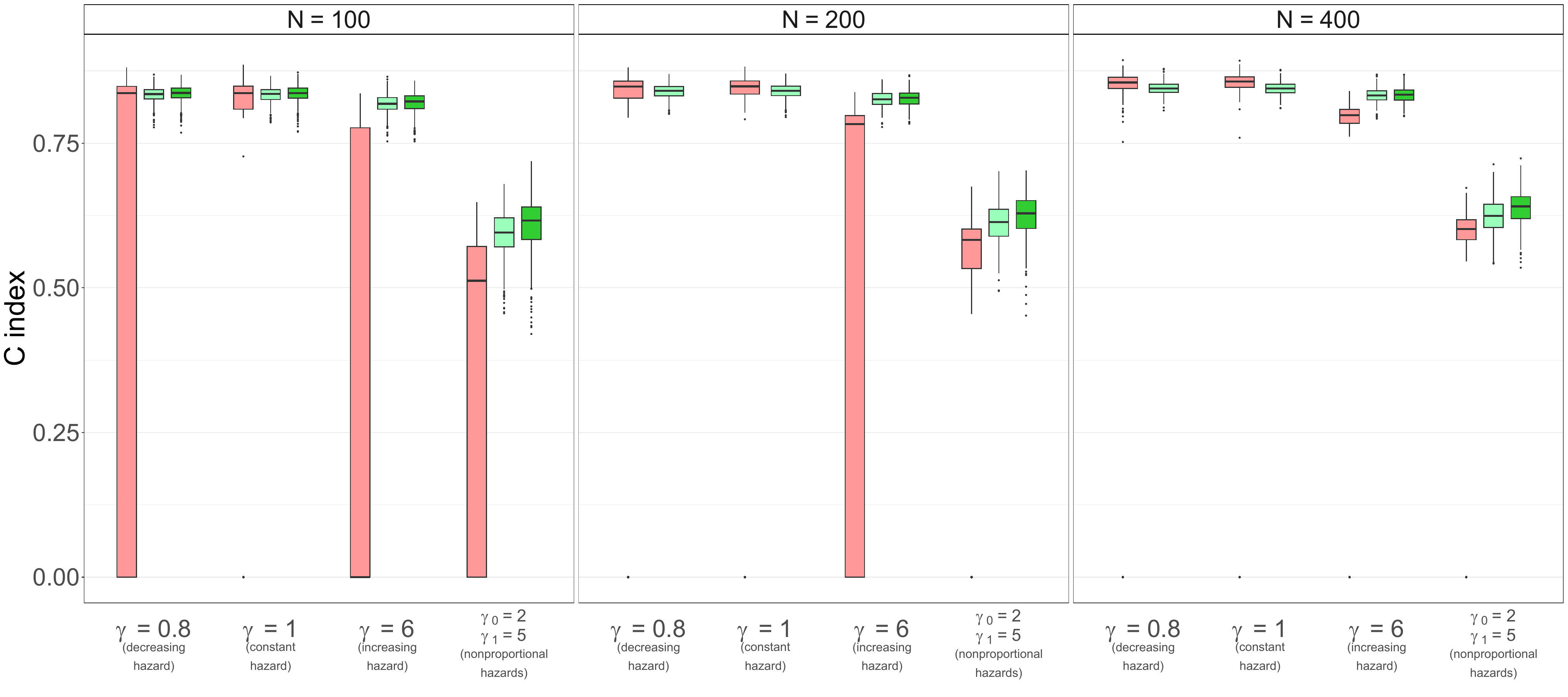}   
							\vspace{-0.7cm}
							\caption[]{\textbf{}}
						\end{minipage}		
						\begin{minipage}{0.005\textwidth} 	
							\includegraphics[scale = 0.325,trim={0cm, 0, 0cm, 0},clip]{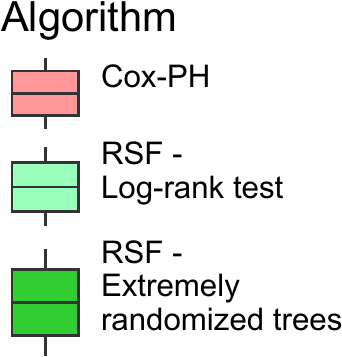}
						\end{minipage}	
					\end{subfigure}			
					
					\vspace{-0.3cm}
					\caption[$C$ index estimates in the simulation scenario assuming 60\% censoring and a treatment effect of $\beta_{\text{\normalfont{treatment}}} = -0.4$ for the RCT in primary biliary cirrhosis (a) and in prostate cancer patients (b).]{{\small  \textbf{$C$ index estimates in the simulation scenario assuming 60\% censoring and a treatment effect of $\beta_{\text{\normalfont{treatment}}} = -0.4$ for the RCT in primary biliary cirrhosis (a) and in prostate cancer patients (b).} Survival times are generated from a Weibull distribution with scale parameters estimated from the respective reference dataset, shape parameters ($\gamma$) vary in order to examine the impact of differing hazards, and the violation of the proportional hazards assumption.
							Results are shown for different total sample sizes $N$.}\\
						{\footnotesize  Abbreviations:  Cox-PH - Cox proportional hazards model, RSF - Random survival forest.}
					} 
					\label{boxplots_cindex_beta0_cens_60}				
				\end{figure}

\begin{figure}[H]
					\begin{subfigure}{\textwidth}
						\begin{minipage}{0.98\textwidth} 	
							\renewcommand\thefigure{}
							\renewcommand{\figurename}{}
							\includegraphics[scale = 0.2, width = 0.995\textwidth,trim={0cm, 0, 0cm, 0},clip]{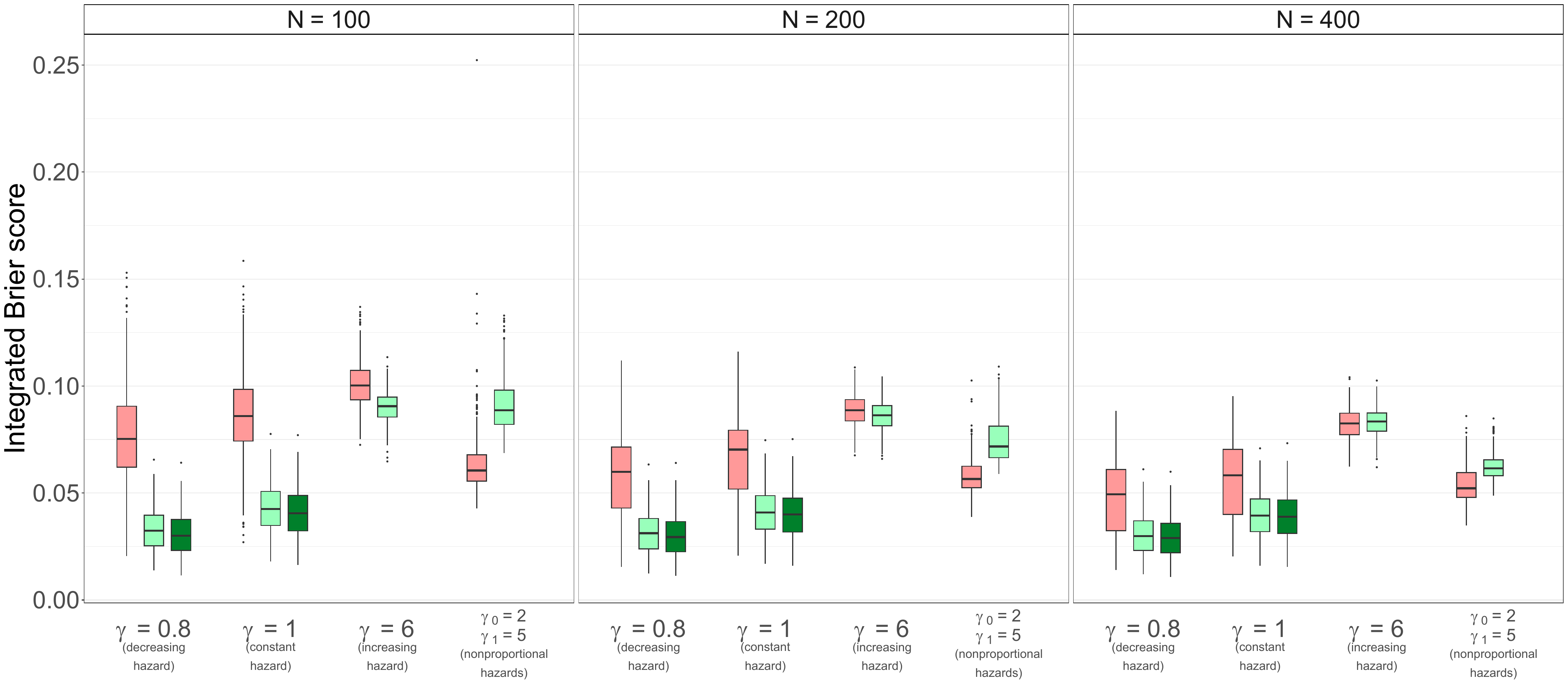}   
							\vspace{-0.7cm}
							\caption[]{\textbf{}}		
						\end{minipage}		
						\begin{minipage}{0.005\textwidth} 	
							\includegraphics[scale = 0.325,trim={0cm, 0, 0cm, 0},clip]{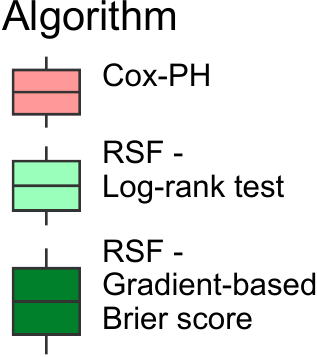}
						\end{minipage}	
					\end{subfigure}		
					\begin{subfigure}{\textwidth}	
						\begin{minipage}{0.98\textwidth} 	
							\renewcommand\thefigure{}
							\renewcommand{\figurename}{}
							\includegraphics[scale = 0.2, width = 0.995\textwidth,trim={0cm, 0, 0cm, 0},clip]{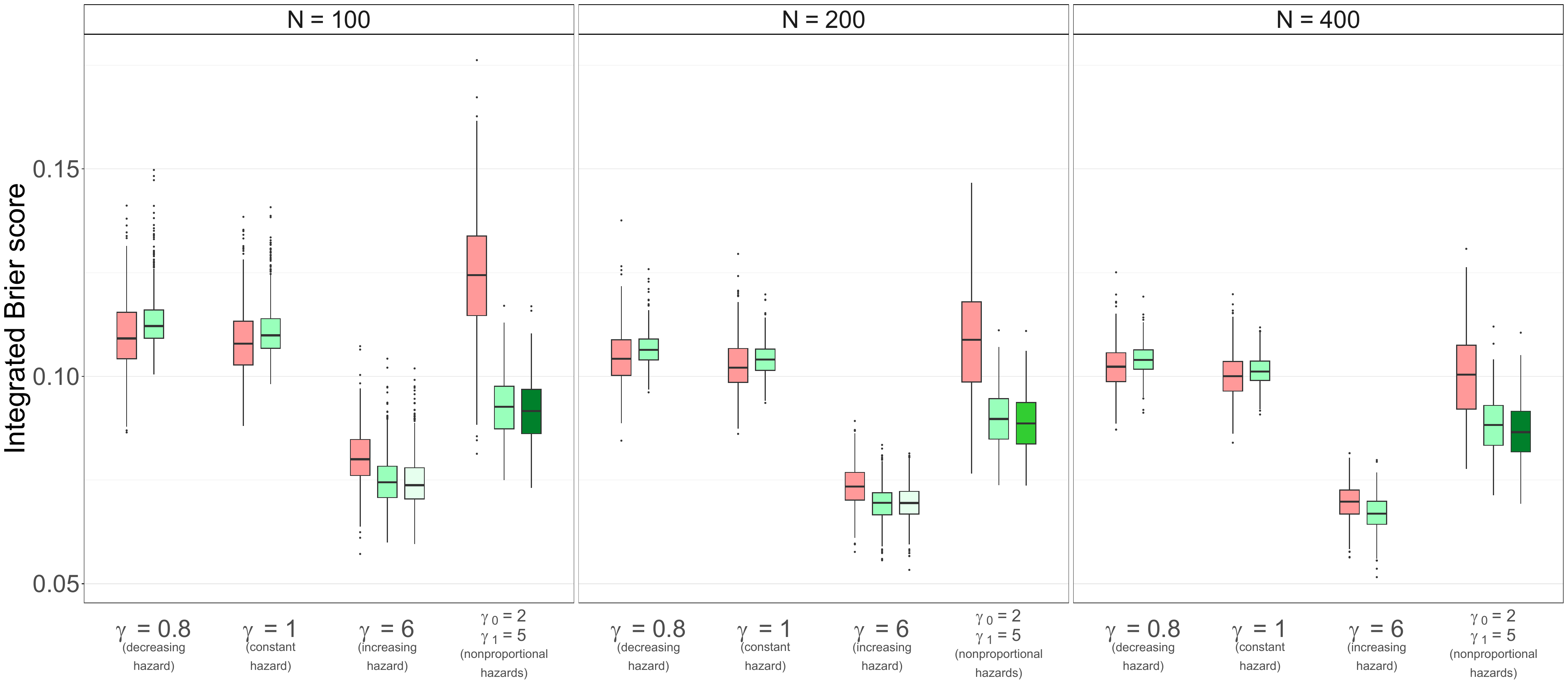}   
							\vspace{-0.7cm}
							\caption[]{\textbf{}}
						\end{minipage}		
						\begin{minipage}{0.005\textwidth} 	
							\includegraphics[scale = 0.325,trim={0cm, 0, 0cm, 0},clip]{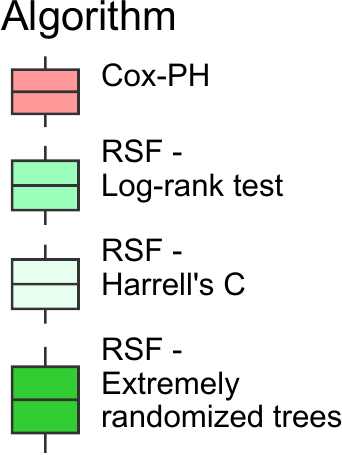}
						\end{minipage}	
					\end{subfigure}	
					\vspace{-0.3cm}
					\caption[Integrated Brier score (IBS) estimates in the simulation scenario assuming 30\% censoring and a treatment effect of $\beta_{\text{\normalfont{treatment}}} = -0.4$ for the RCT in primary biliary cirrhosis (a) and in prostate cancer patients (b).]{{\small{\textbf{Integrated Brier score (IBS) estimates in the simulation scenario assuming 30\% censoring and a treatment effect of $\beta_{\text{\normalfont{treatment}}} = -0.4$ for the RCT in primary biliary cirrhosis (a) and in prostate cancer patients (b).}} Survival times are generated from a Weibull distribution with scale parameters estimated from the respective reference dataset, shape parameters ($\gamma$) vary in order to examine the impact of differing hazards, and the violation of the proportional hazards assumption.
							Results are shown for different total sample sizes $N$.}\\
						{\footnotesize  Abbreviations:  Cox-PH - Cox proportional hazards model, RSF - Random survival forest.}
					} 
					\label{boxplots_ibs_beta0_cens_30}				
				\end{figure}

\begin{figure}[H]
					\begin{subfigure}{\textwidth}
						\begin{minipage}{0.98\textwidth} 	
							\renewcommand\thefigure{}
							\renewcommand{\figurename}{}
							\includegraphics[scale = 0.2, width = 0.995\textwidth,trim={0cm, 0, 0cm, 0},clip]{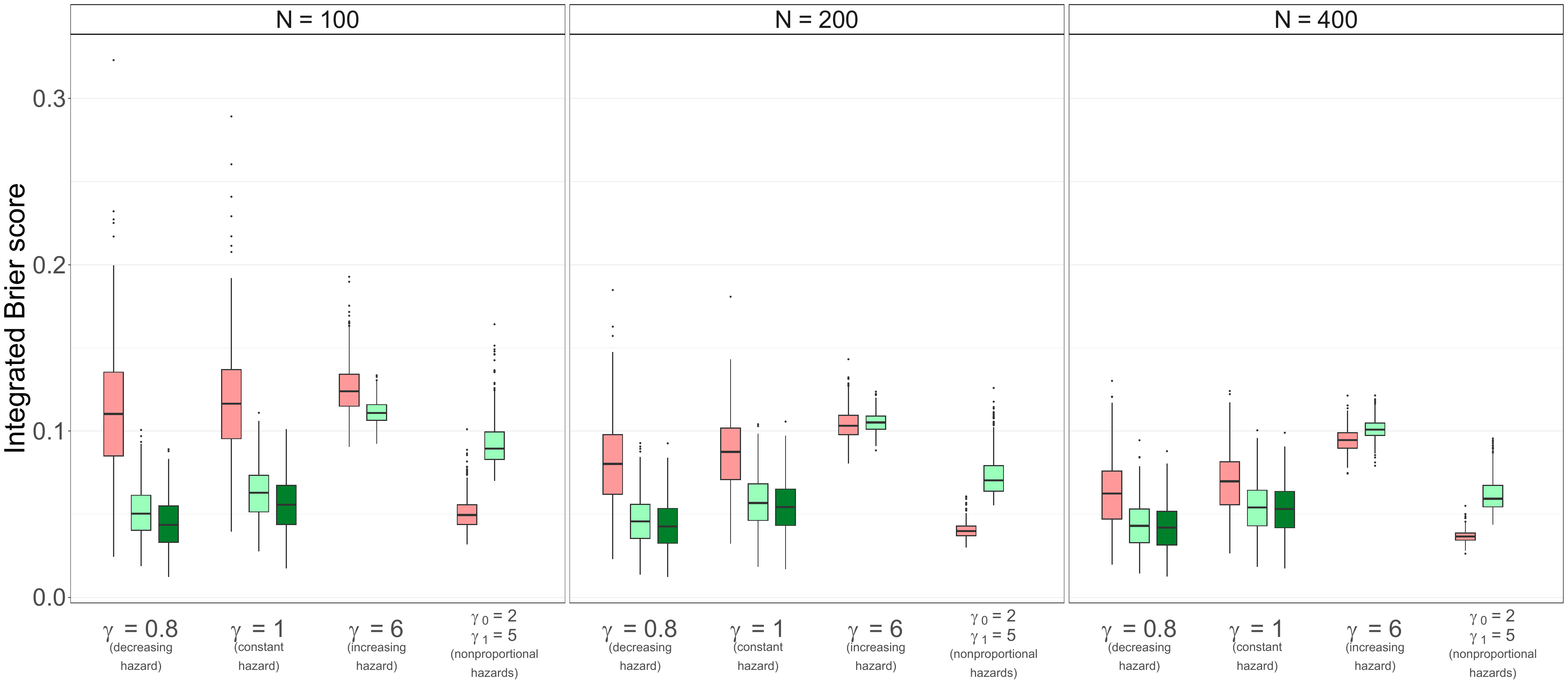}   
							\vspace{-0.7cm}
							\caption[]{\textbf{}}		
						\end{minipage}		
						\begin{minipage}{0.005\textwidth} 	
							\includegraphics[scale = 0.325,trim={0cm, 0, 0cm, 0},clip]{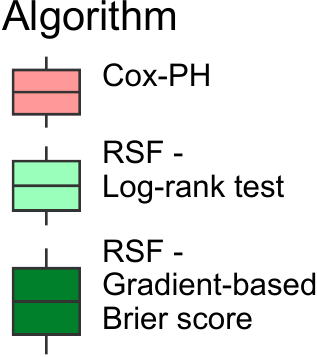}
						\end{minipage}	
					\end{subfigure}		
					\begin{subfigure}{\textwidth}	
						\begin{minipage}{0.98\textwidth} 	
							\renewcommand\thefigure{}
							\renewcommand{\figurename}{}
							\includegraphics[scale = 0.2, width = 0.995\textwidth,trim={0cm, 0, 0cm, 0},clip]{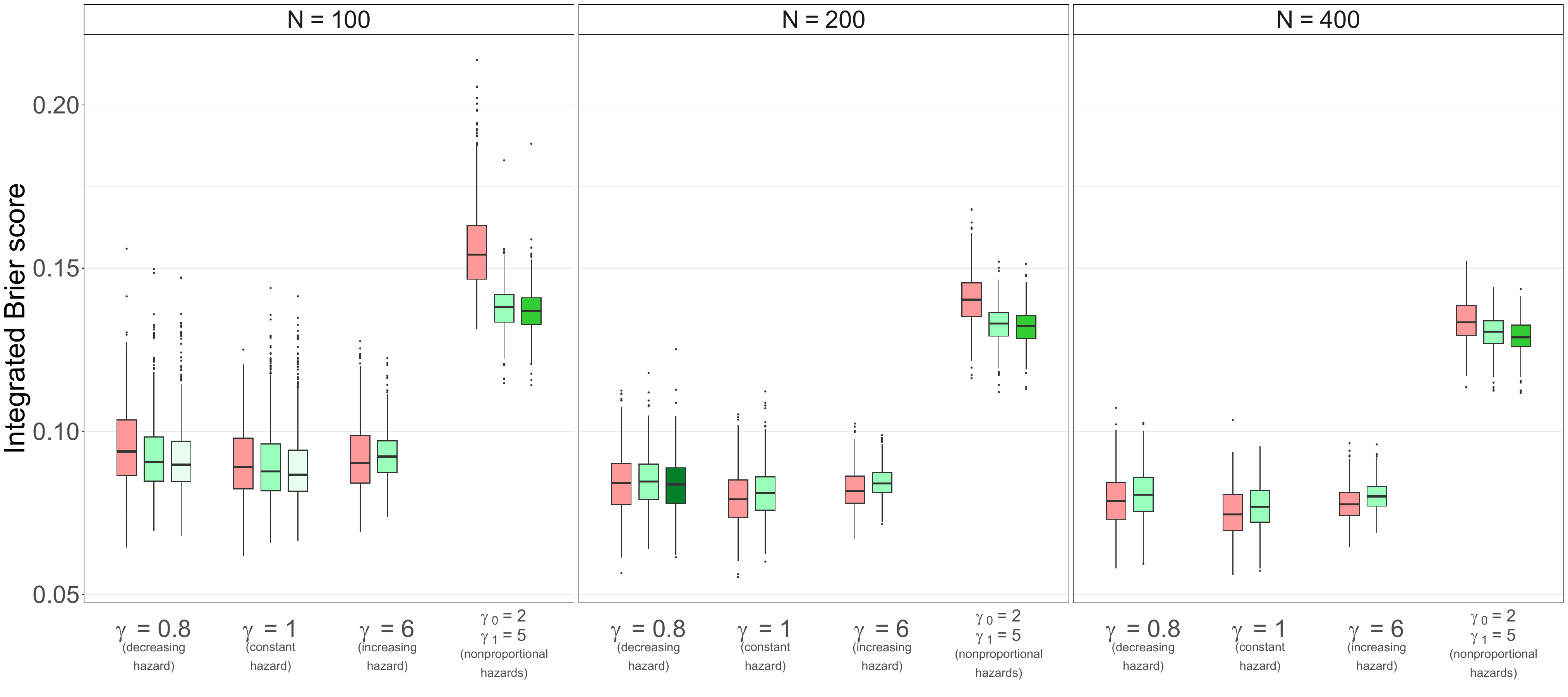}   
							\vspace{-0.7cm}
							\caption[]{\textbf{}}
						\end{minipage}		
						\begin{minipage}{0.005\textwidth} 	
							\includegraphics[scale = 0.325,trim={0cm, 0, 0cm, 0},clip]{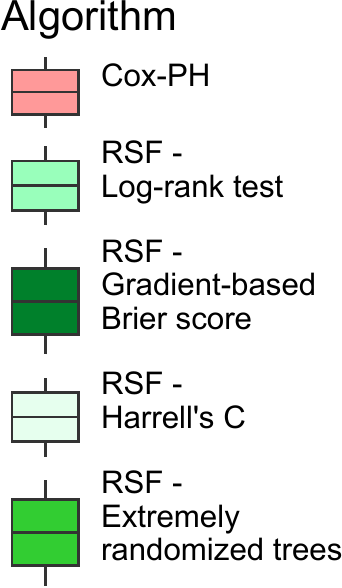}
						\end{minipage}	
					\end{subfigure}	
					\vspace{-0.3cm}
					\caption[Integrated Brier score (IBS) estimates in the simulation scenario assuming 60\% censoring and a treatment effect of $\beta_{\text{\normalfont{treatment}}} = -0.4$ for the RCT in primary biliary cirrhosis (a) and in prostate cancer patients (b).]{{\small{\textbf{Integrated Brier score (IBS) estimates in the simulation scenario assuming 60\% censoring and a treatment effect of $\beta_{\text{\normalfont{treatment}}} = -0.4$ for the RCT in primary biliary cirrhosis (a) and in prostate cancer patients (b).}} Survival times are generated from a Weibull distribution with scale parameters estimated from the respective reference dataset, shape parameters ($\gamma$) vary in order to examine the impact of differing hazards, and the violation of the proportional hazards assumption.
							Results are shown for different total sample sizes $N$.}\\
						{\footnotesize  Abbreviations:  Cox-PH - Cox proportional hazards model, RSF - Random survival forest.}
					} 
					\label{boxplots_ibs_beta0_cens_60}				
				\end{figure}

\begin{figure}[H]
					\begin{minipage}{.45\textwidth}
						\includegraphics[scale = 0.45,trim={0 0 3.5cm 0},clip]{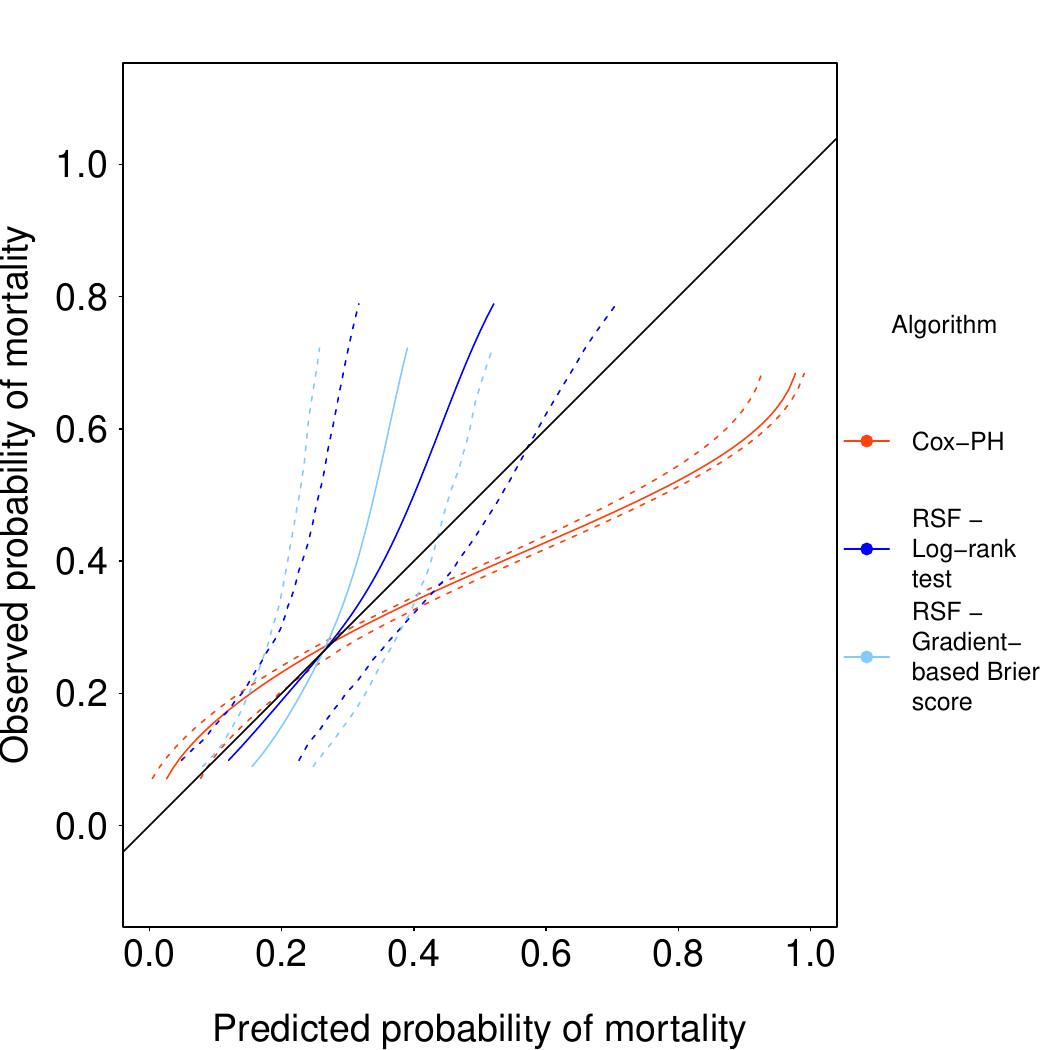}
						\begin{center}
							(a)
						\end{center}
						\includegraphics[scale = 0.45,trim={0 0 3.5cm 0},clip]{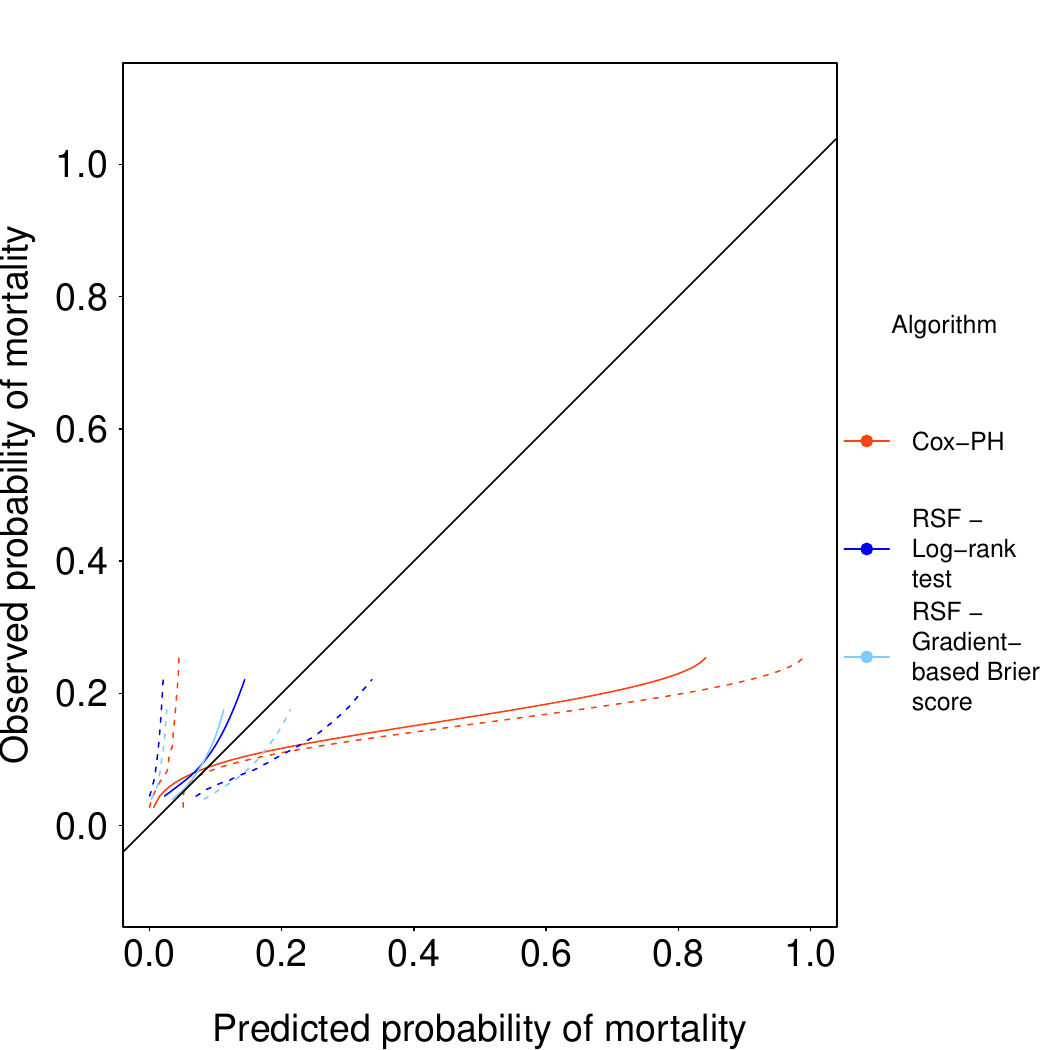}
						\begin{center}
							(c)
						\end{center}
					\end{minipage}
					\begin{minipage}{.029\textwidth}
					\end{minipage}			
					\begin{minipage}{.52\textwidth}
						\includegraphics[scale = 0.45,trim={0.7cm 0 0 0},clip]{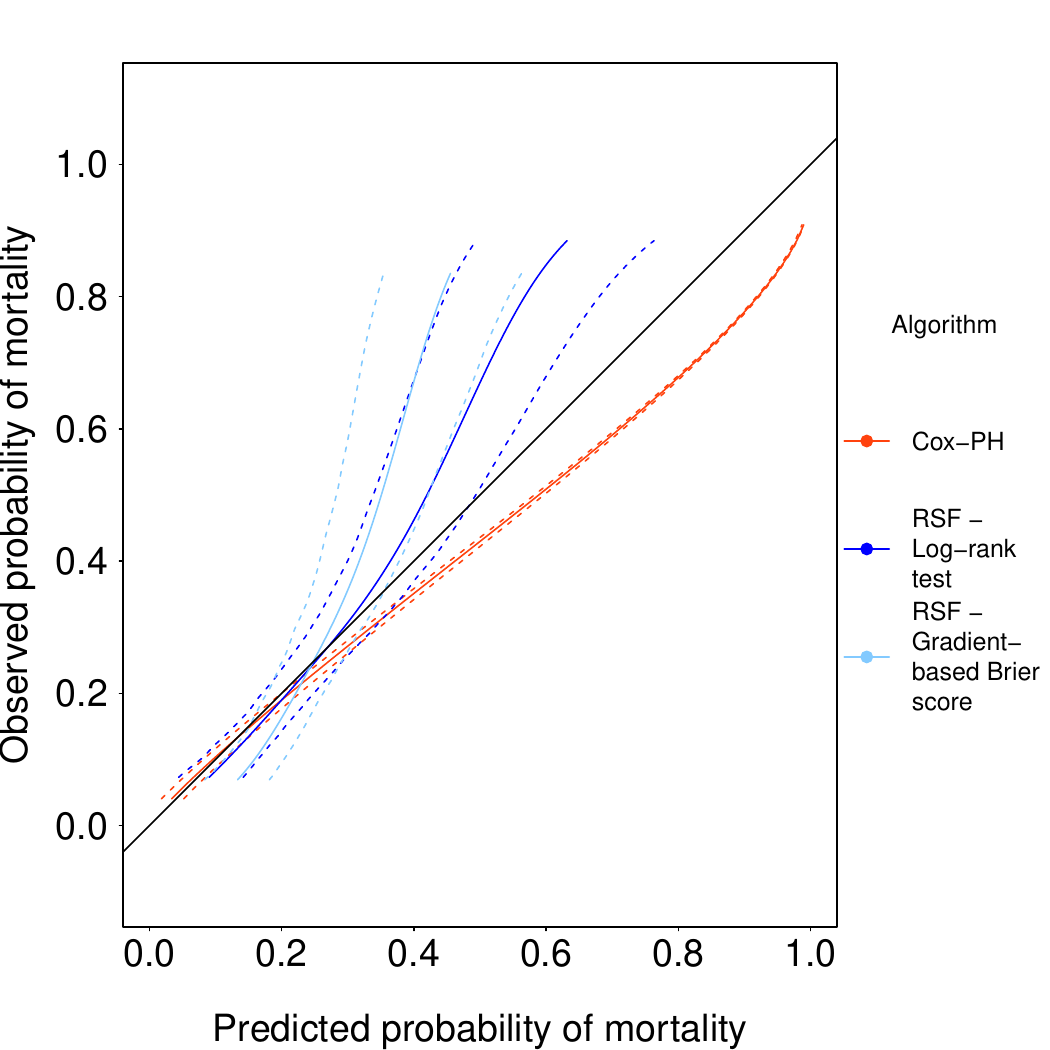}
						\begin{center}
							(b)
						\end{center}
						\includegraphics[scale = 0.45,trim={0.7cm 0 0 0},clip]{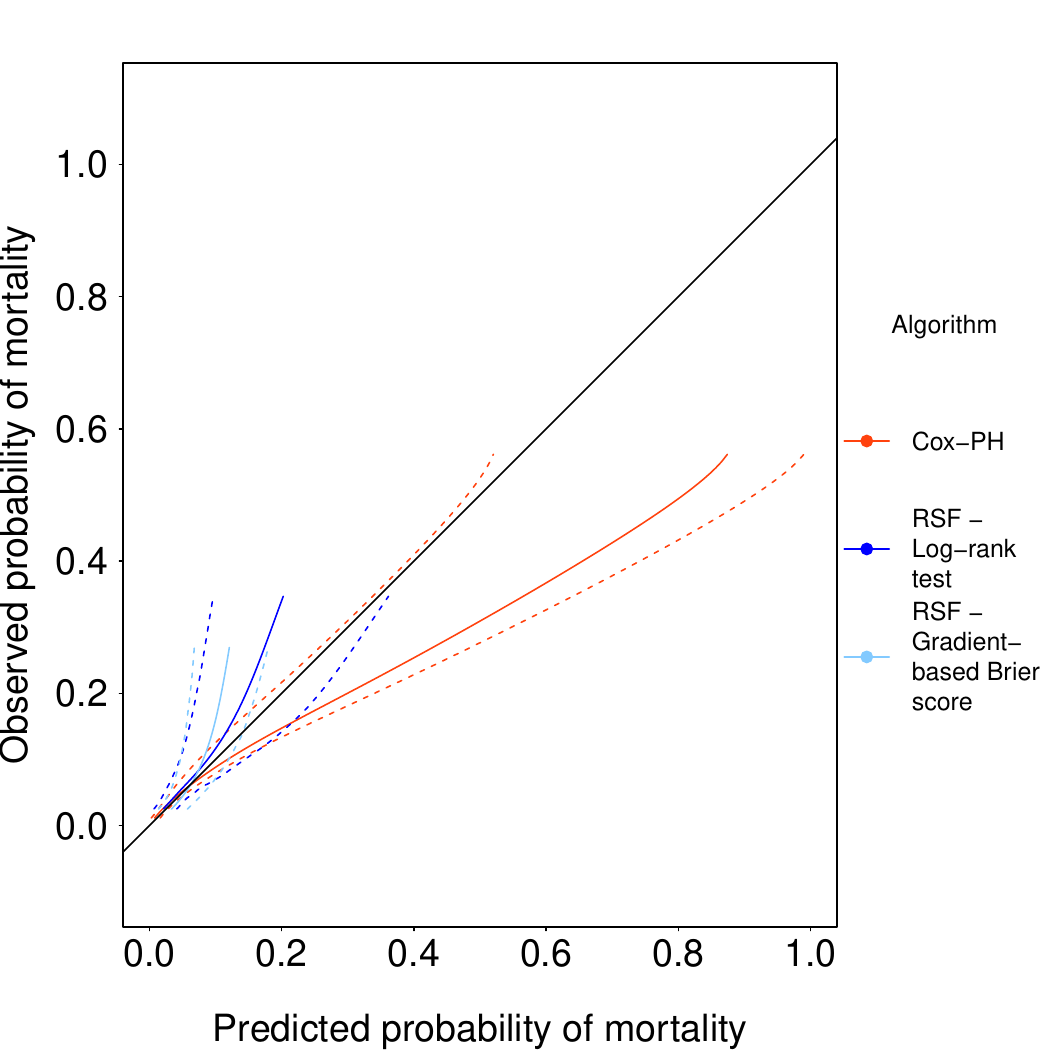}			
						\begin{center}
							(d)
						\end{center}				
					\end{minipage}
					\captionof{figure}[Calibration curves at the median survival time for the data without treatment-covariate interactions (primary biliary cirrhosis dataset) for a proportional hazard setting.]{\linespread{1}\selectfont \small{\textbf{Calibration curves for a proportional hazards scenario (primary biliary cirrhosis dataset).}  \underline{Calibration curves at the median (50\% quantile)  survival time} for a \underline{proportional hazards} setting (Weibull survival time distribution W($\lambda = 2241.74, \gamma = 1$)), $\beta_{\text{treatment}} = -0.4$, and $n_{\text{sim}} = 500$ simulated datasets  based on data  \underline{without treatment-covariate interactions (primary biliary cirrhosis dataset)}. The solid line represents the mean calibration curve, the outer dotted lines represent the 2.5th and 97.5th percentile of the calibration curve. The black diagonal line corresponds to perfect calibration.\\
							(a) 30\% censoring, $N = 100$, (b) 30\% censoring, $N = 400$, \\(c) 60\% censoring, $N = 100$, (d) 60\% censoring, $N = 400$.\\  }  			
						{  \footnotesize  Abbreviations: Cox-PH - Cox proportional hazards model, RSF - Random survival forest.}		
					}
					
					\label{Mayo_calib_treat_04_ph}		
				\end{figure}

\begin{figure}[H]
					\begin{minipage}{.45\textwidth}
						\includegraphics[scale = 0.45,trim={0 0 3.5cm 0},clip]{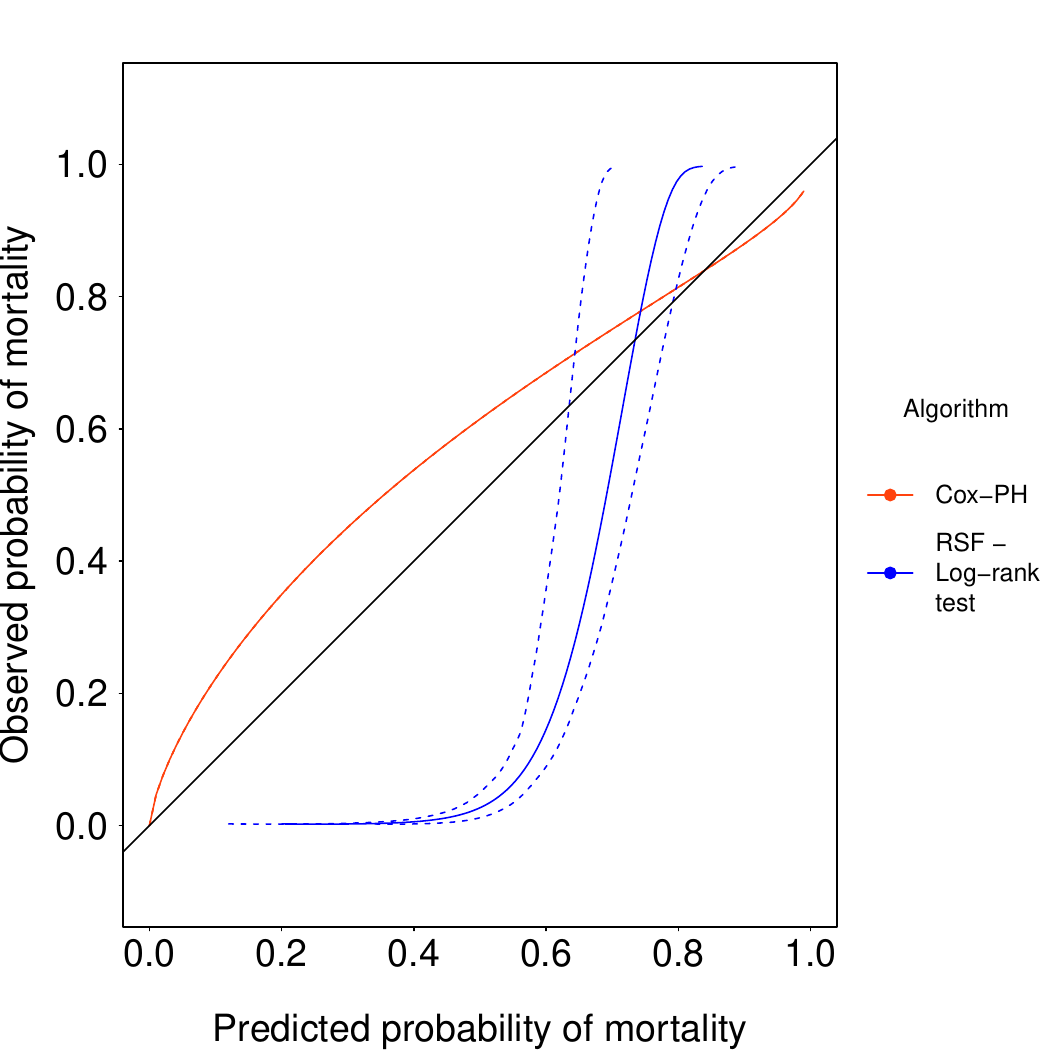}
						\begin{center}
							(a)
						\end{center}
						\includegraphics[scale = 0.45,trim={0 0 3.5cm 0},clip]{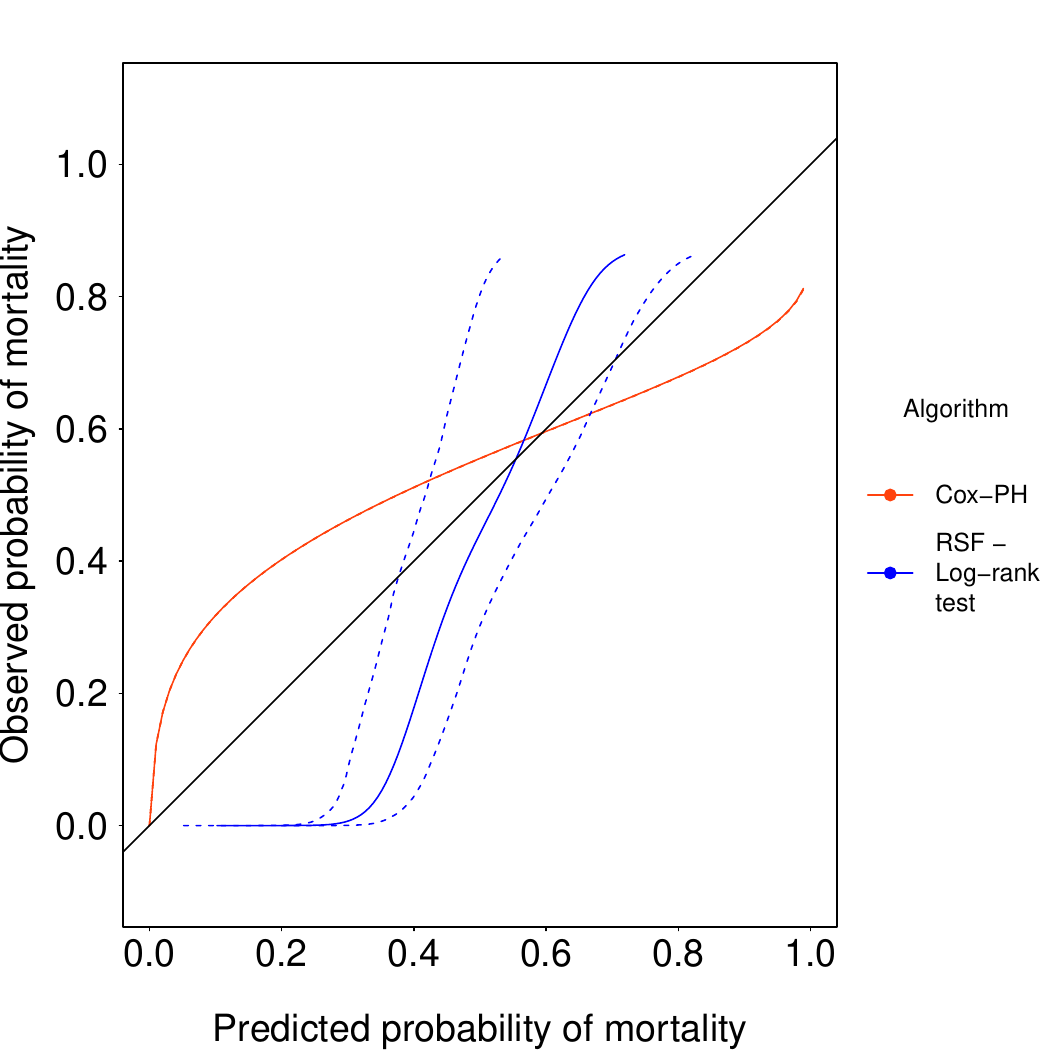}
						\begin{center}
							(c)
						\end{center}
					\end{minipage}
					\begin{minipage}{.029\textwidth}
					\end{minipage}
					\begin{minipage}{.52\textwidth}
						\includegraphics[scale = 0.45,trim={0.7cm 0 0 0},clip]{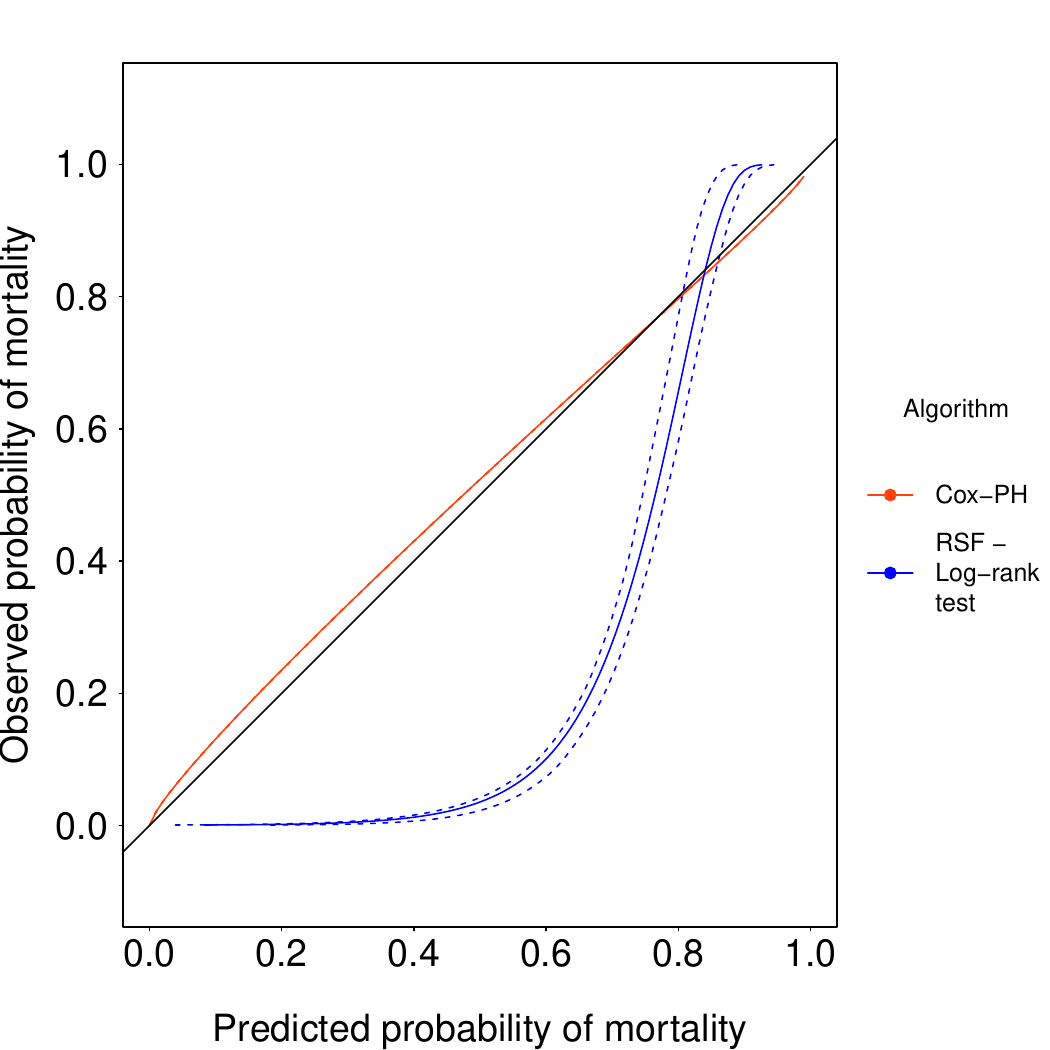}
						\begin{center}
							(b)
						\end{center}
						\includegraphics[scale = 0.45,trim={0.7cm 0 0 0},clip]{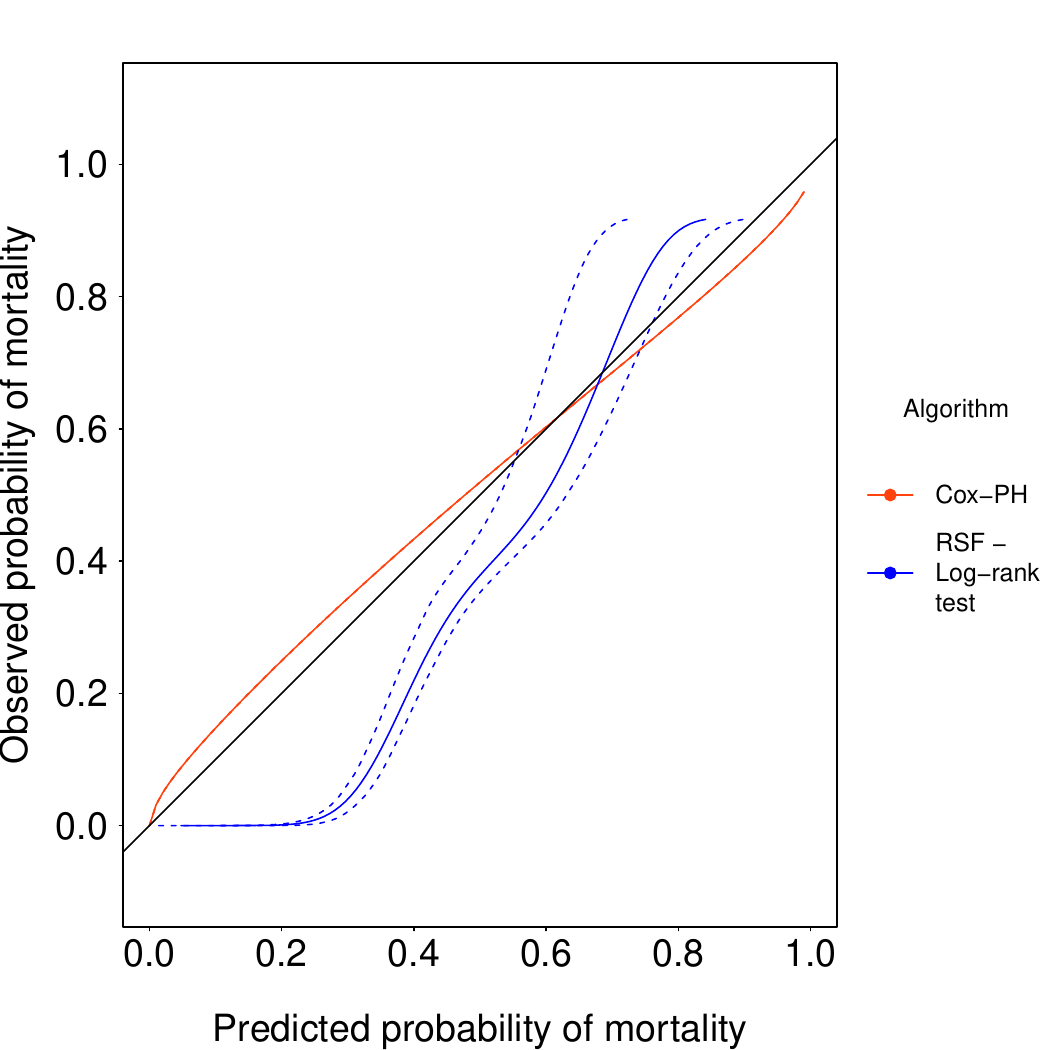}			
						\begin{center}
							(d)
						\end{center}				
					\end{minipage}
					\vspace{0.1cm}
					\captionof{figure}[Calibration curves at the median survival time for the data without treatment-covariate interactions (primary biliary cirrhosis dataset) for a nonproportional hazard setting.]{\linespread{1}\selectfont \small{\textbf{Calibration curves for a nonproportional hazards setting (primary biliary cirrhosis dataset).}  \underline{Calibration curves at the median (50\% quantile)  survival time} for a \underline{nonproportional hazards} setting (Weibull survival time distribution W($ \lambda = 2241.74, \gamma \in \{2,5\}$)), $\beta_{\text{treatment}} = -0.4$, and $n_{\text{sim}} = 500$ simulated datasets  based on data  \underline{without treatment-covariate interactions (primary biliary cirrhosis dataset)}. The solid line represents the mean calibration curve, the outer dotted lines represent the 2.5th and 97.5th percentile of the calibration curve. The black diagonal line corresponds to perfect calibration.\\
							(a) 30\% censoring, $N = 100$, (b) 30\% censoring, $N = 400$, \\(c) 60\% censoring, $N = 100$, (d) 60\% censoring, $N = 400$.\\   }  			
						{  \footnotesize  Abbreviations: Cox-PH - Cox proportional hazards model, RSF - Random survival forest.}		
					}
					
					\label{Mayo_calib_treat_04_nonph}	
				\end{figure}

\begin{figure}[H]
					\begin{minipage}{.45\textwidth}
						\includegraphics[scale = 0.45,trim={0 0 4.2cm 0},clip]{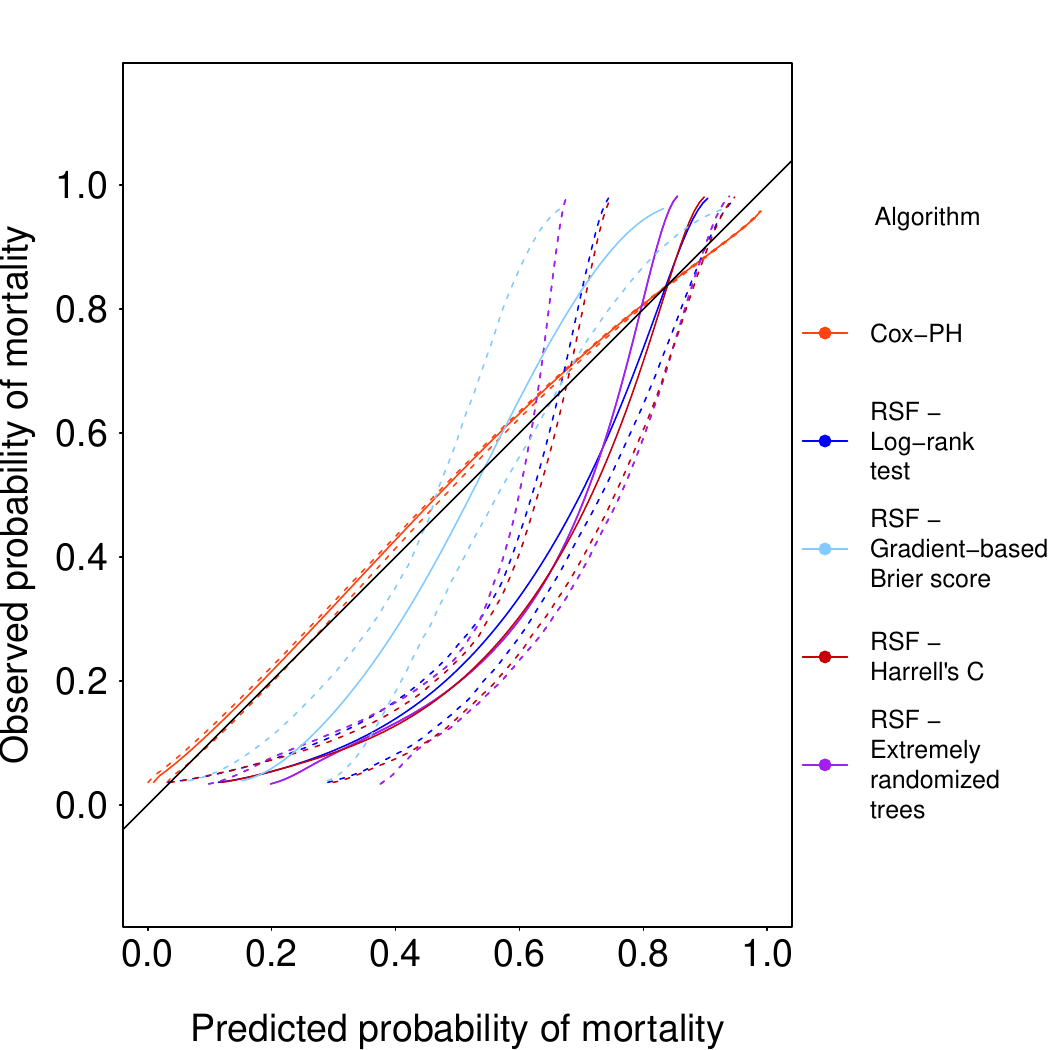}
						\begin{center}
							(a)
						\end{center}
						\includegraphics[scale = 0.45,trim={0 0 4.2cm 0},clip]{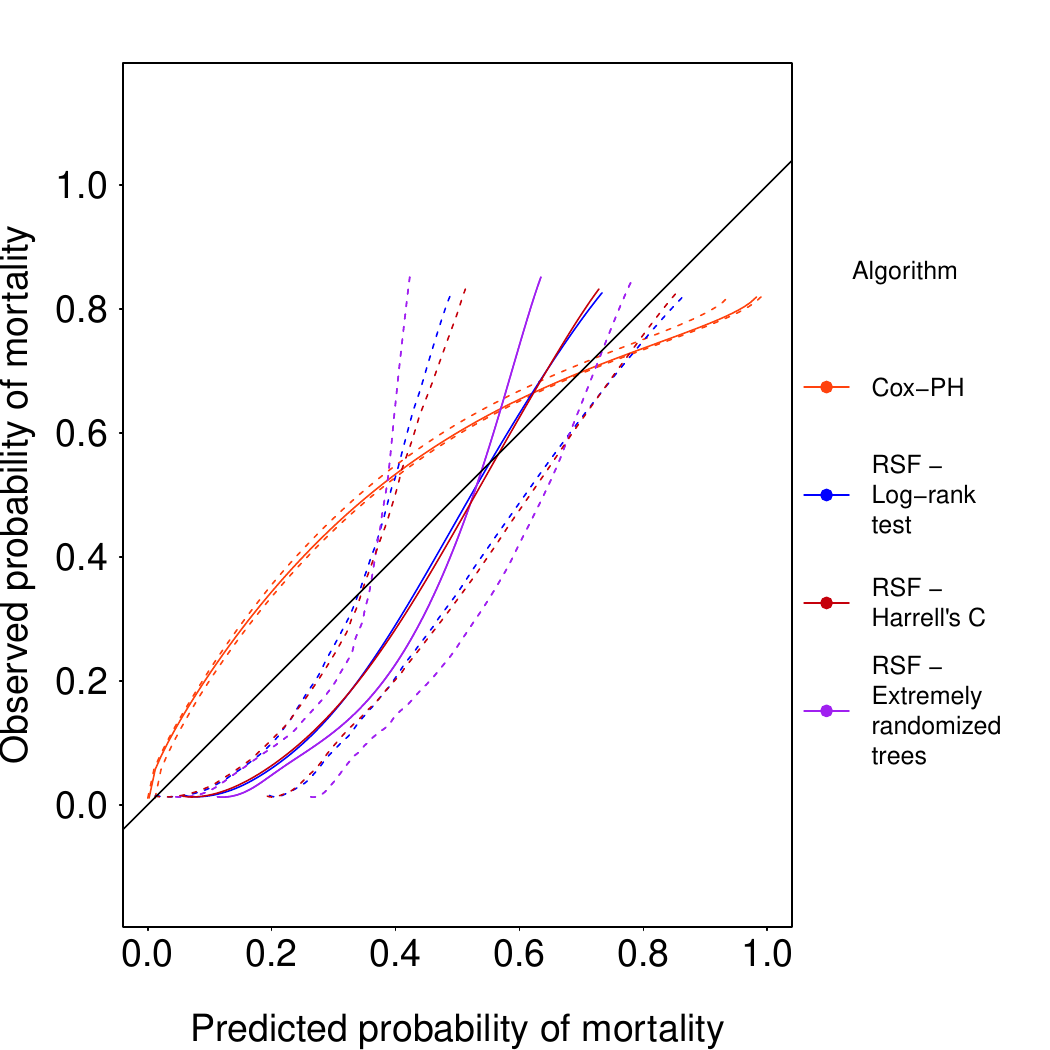}
						\begin{center}
							(c)
						\end{center}
					\end{minipage}
					\begin{minipage}{.029\textwidth}
					\end{minipage}
					\begin{minipage}{.52\textwidth}
						\includegraphics[scale = 0.45,trim={0.7cm 0 0 0},clip]{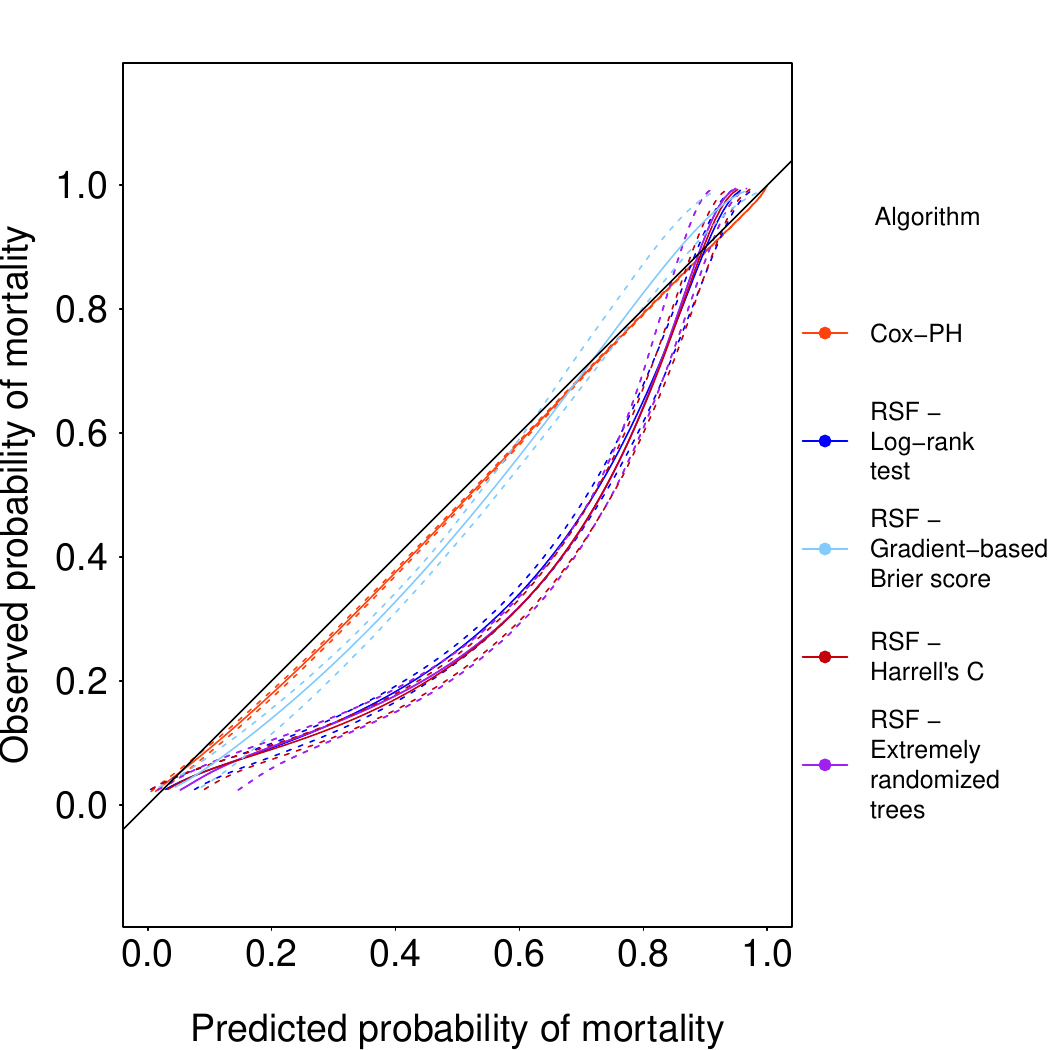}
						\begin{center}
							(b)
						\end{center}
						\includegraphics[scale = 0.45,trim={0.7cm 0 0 0},clip]{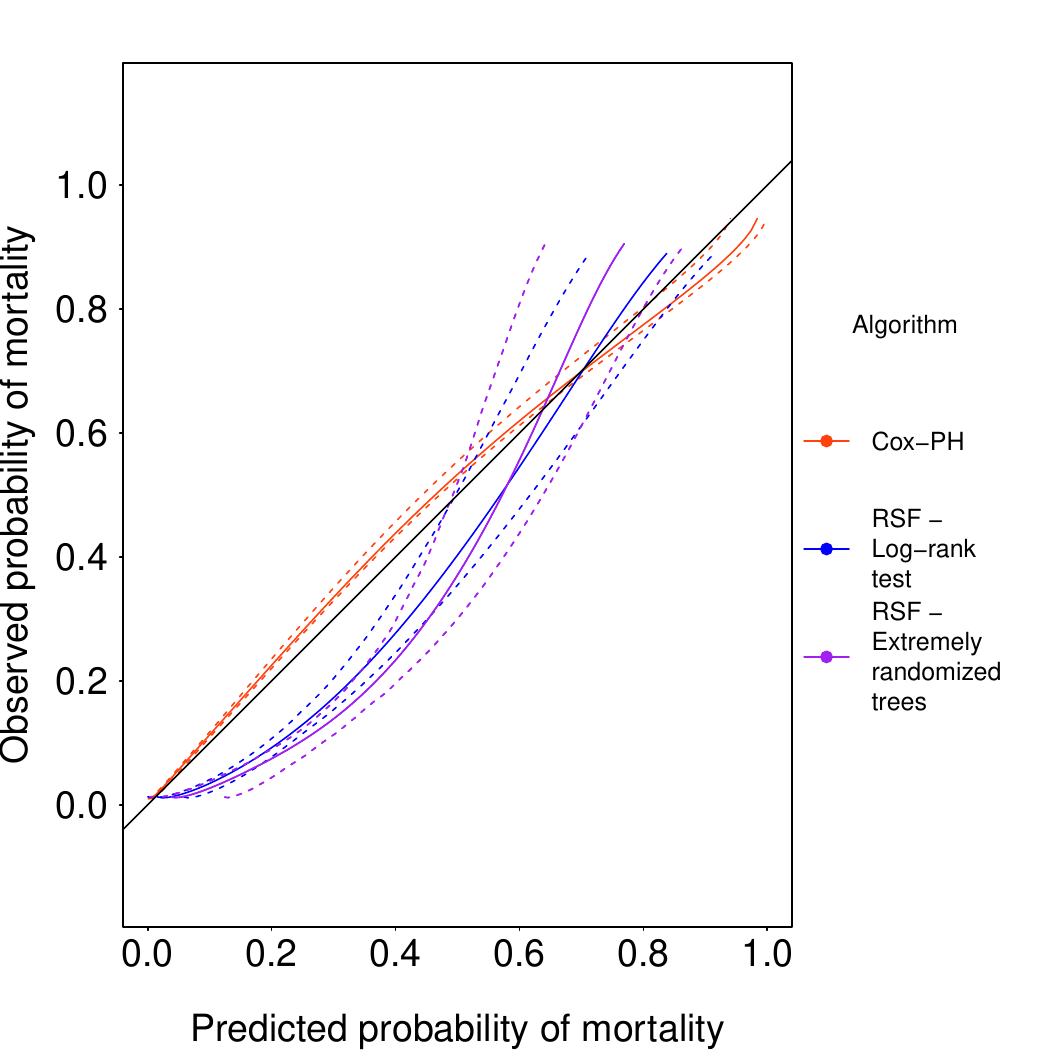}			
						\begin{center}
							(d)
						\end{center}				
					\end{minipage}
					\captionof{figure}[Calibration curves at the median survival time for the data with three treatment-covariate interactions (prostate cancer dataset) for a proportional hazard setting.]{\linespread{1}\selectfont \small{\textbf{Calibration curves for a proportional hazards setting (prostate cancer dataset).}  \underline{Calibration curves at the median (50\% quantile)  survival time} for a \underline{proportional hazards} setting (Weibull survival time distribution W($\lambda = 2241.74, \gamma = 1$)), $\beta_{\text{treatment}} = -0.4$, and $n_{\text{sim}} = 500$ simulated datasets  based on data  \underline{with three treatment-covariate interactions (prostate cancer dataset)}. The solid line represents the mean calibration curve, the outer dotted lines represent the 2.5th and 97.5th percentile of the calibration curve. The black diagonal line corresponds to perfect calibration.\\
							(a) 30\% censoring, $N = 100$, (b) 30\% censoring, $N = 400$, \\(c) 60\% censoring, $N = 100$, (d) 60\% censoring, $N = 400$.\\ }  			
						{  \footnotesize  Abbreviations: Cox-PH - Cox proportional hazards model, RSF - Random survival forest.}		
					}
					
					\label{Byar_calib_treat_04_ph}		
				\end{figure}

\begin{figure}[H]
					\begin{minipage}{.45\textwidth}
						\includegraphics[scale = 0.45,trim={0 0 4.2cm 0},clip]{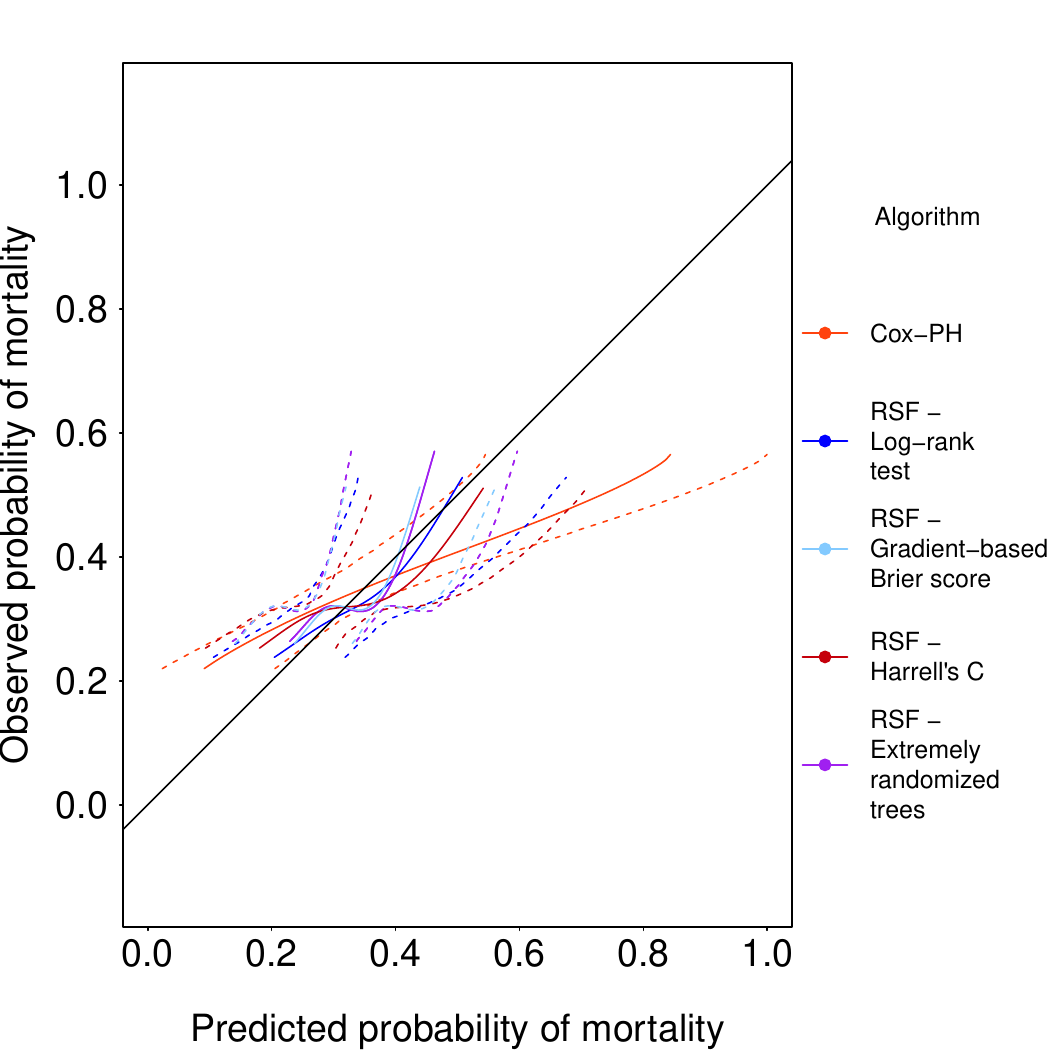}
						\begin{center}
							(a)
						\end{center}
						\includegraphics[scale = 0.45,trim={0 0 4.2cm 0},clip]{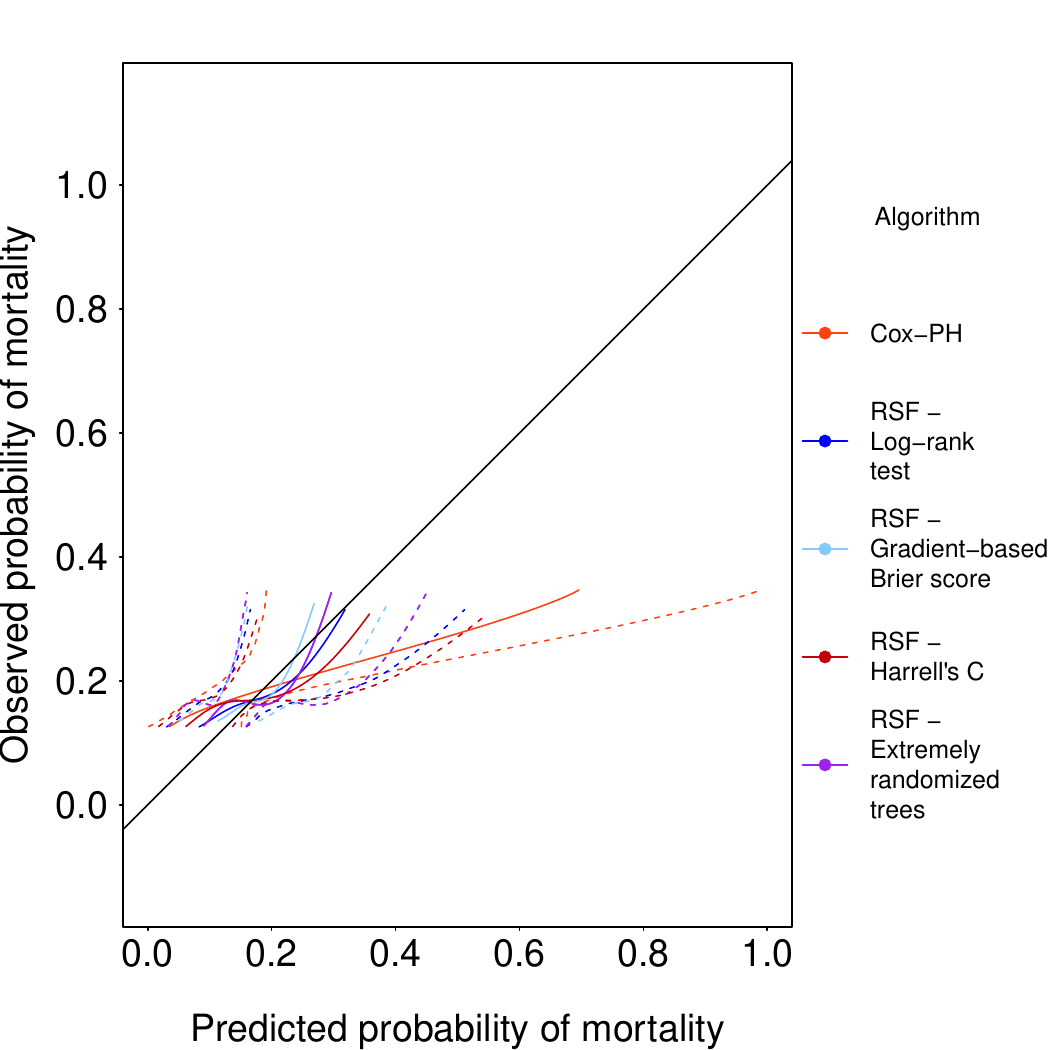}
						\begin{center}
							(c)
						\end{center}
					\end{minipage}
					\begin{minipage}{.029\textwidth}
					\end{minipage}
					\begin{minipage}{.52\textwidth}
						\includegraphics[scale = 0.45,trim={0.7cm 0 0 0},clip]{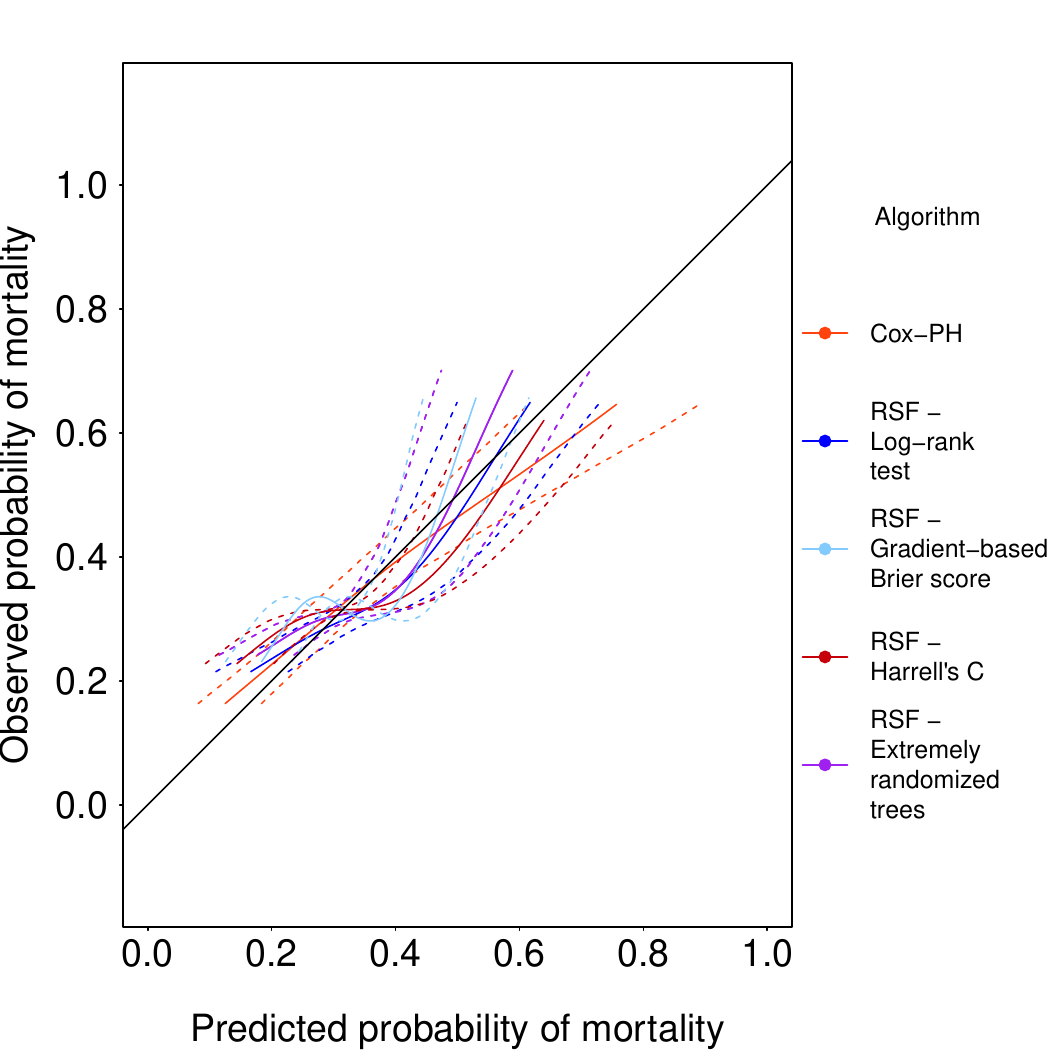}
						\begin{center}
							(b)
						\end{center}
						\includegraphics[scale = 0.45,trim={0.7cm 0 0 0},clip]{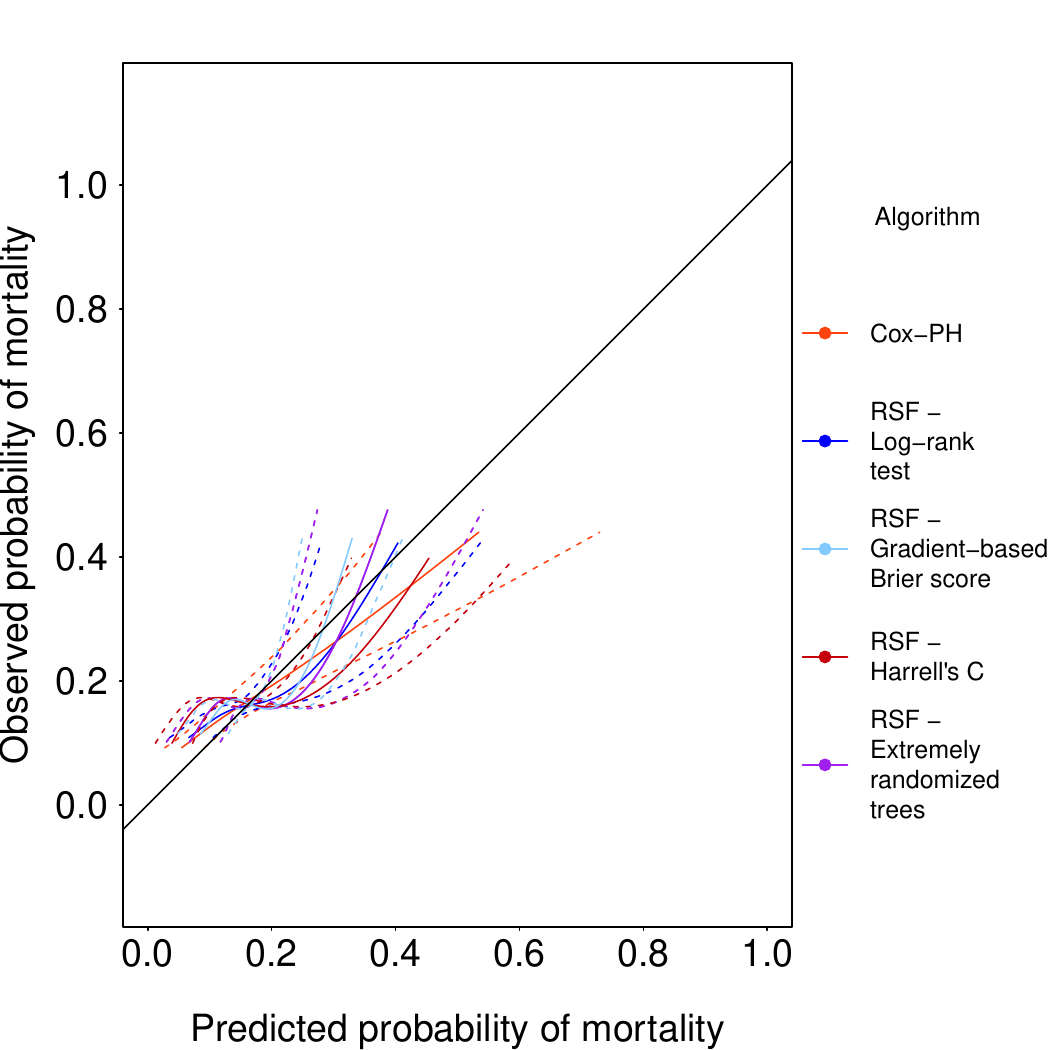}			
						\begin{center}
							(d)
						\end{center}				
					\end{minipage}
					\captionof{figure}[Calibration curves at the median survival time for the data with three treatment-covariate interactions (prostate cancer dataset) for a nonproportional hazard setting.]{\linespread{1}\selectfont \small{\textbf{Calibration curves for a nonproportional hazards setting (prostate cancer dataset).}  \underline{Calibration curves at the median (50\% quantile)  survival time} for a nonproportional hazard setting (Weibull survival time distribution W($\lambda = 39.2, \gamma \in \{2,5\}$)), $\beta_{\text{treatment}} = -0.4$, and $n_{\text{sim}} = 500$ simulated datasets  based on data  \underline{with three treatment-covariate interactions (prostate cancer dataset)}. The solid line represents the mean calibration curve, the outer dotted lines represent the 2.5th and 97.5th percentile of the calibration curve. The black diagonal line corresponds to perfect calibration.\\
							(a) 30\% censoring, $N = 100$, (b) 30\% censoring, $N = 400$, \\(c) 60\% censoring, $N = 100$, (d) 60\% censoring, $N = 400$.\\ }  			
						{  \footnotesize  Abbreviations: Cox-PH - Cox proportional hazards model, RSF - Random survival forest.}		
					}
					
					\label{Byar_calib_treat_04_nonph}	
				\end{figure}

\begin{figure}[H]
					\begin{minipage}{.45\textwidth}
						\includegraphics[scale = 0.235,trim={0 0 10.5cm 0},clip]{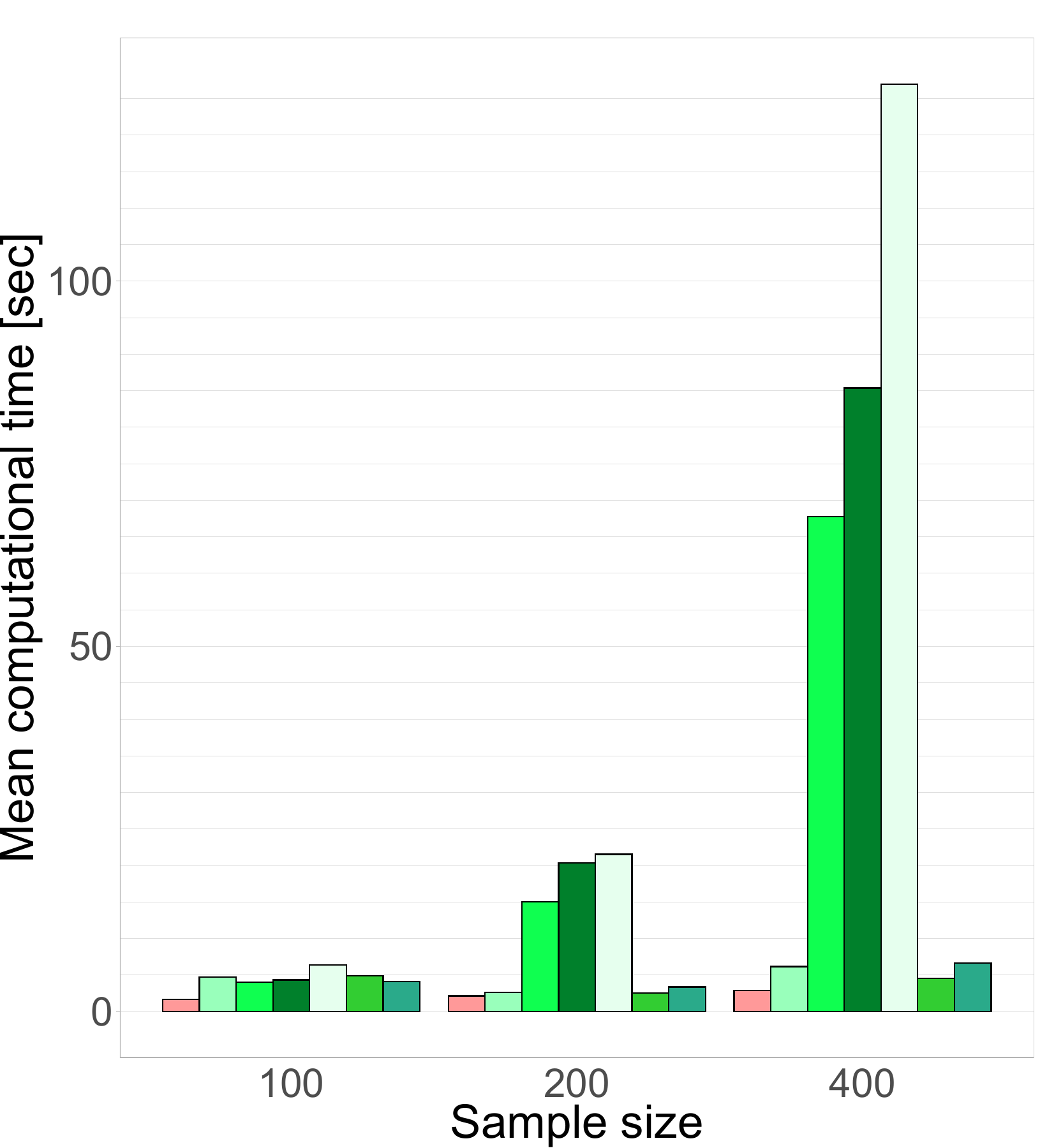}
					\end{minipage}
					\begin{minipage}{.029\textwidth}
					\end{minipage}
					\begin{minipage}{.52\textwidth}
						\vspace{0.2cm}
						\includegraphics[scale = 0.23,trim={0 0 0 0},clip]{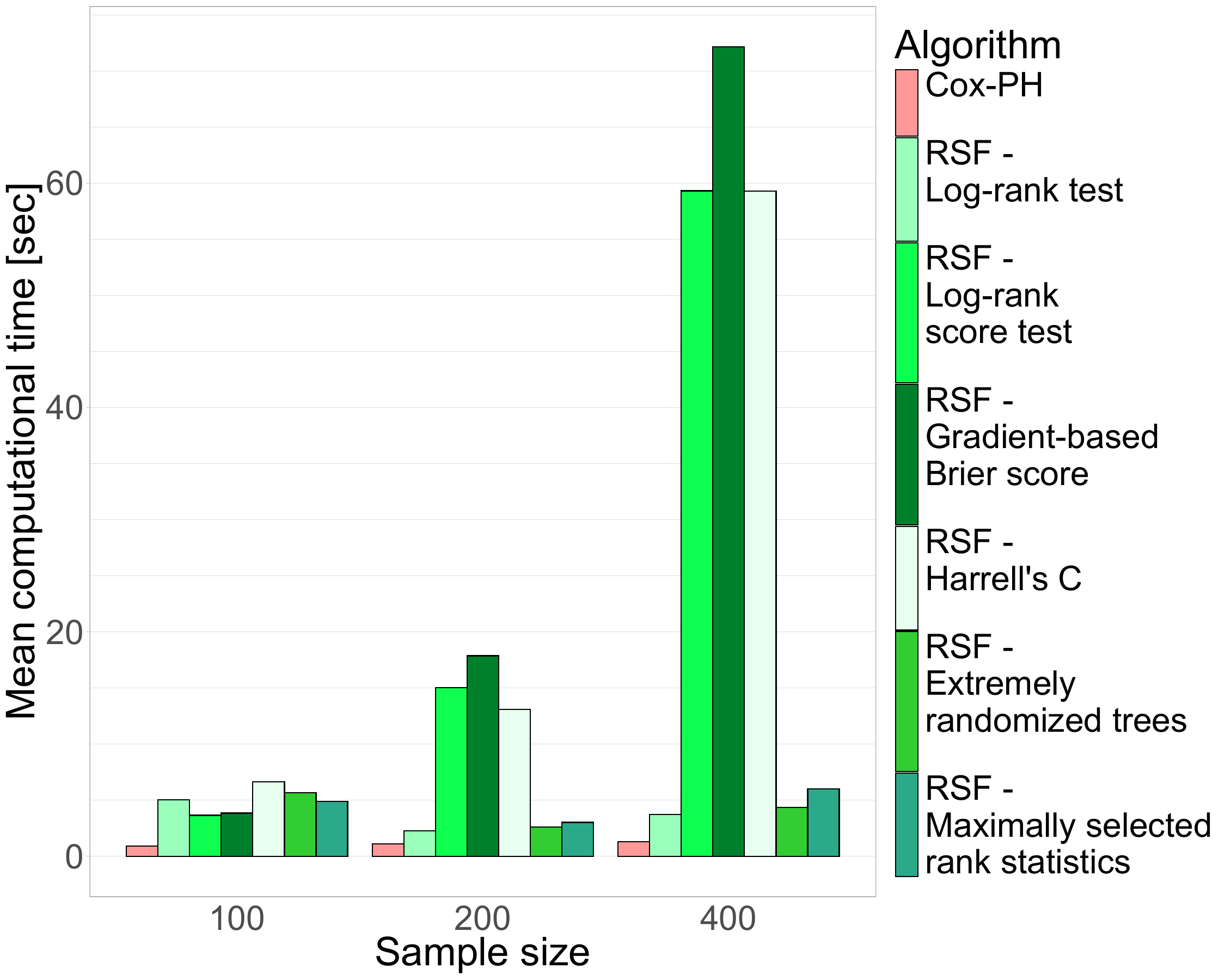}				
					\end{minipage}
					\vspace{0.1cm}
					\captionof{figure}[Mean computational times for the RCT data without treatment-covariate interactions (primary biliary cirrhosis dataset), and for the RCT data with three treatment-covariate interactions (prostate cancer dataset).]{\linespread{1}\selectfont \small{\textbf{Mean computational times for the RCT data without treatment-covariate interactions (primary biliary cirrhosis dataset, left), and for the RCT data with three treatment-covariate interactions (prostate cancer dataset, right).} \\ }  			
						{  \footnotesize  Abbreviations: Cox-PH - Cox proportional hazards model, RSF - Random survival forest.}		
					}
					
					\label{comp_times}	
				\end{figure}

				In this section, the simulation study results for one of the treatment effects considered in the simulation study ($\beta_{\text{treatment}} = -0.4$) are presented and discussed. The results for other values of the treatment effect ($\beta_{\text{treatment}} \in \{0, 0.8\}$) are similar and can be found in the Supplementary Material  C. 
				Moreover, only the results for the algorithms that are of most interest are shown, which are the Cox model and the RSF using the standard log-rank test splitting rule. Additionally, the results of the RSF based on other splitting rules are shown if they outperform these two methods with respect to the median result. Only the best performing one among them is shown in case there are multiple better performing alternatives. Results for the remaining RSF splitting rules are shown in the Supplementary Material  C. \\				
				The $C$ index estimates, which correspond to the models' discriminative performance, are shown in Figure \ref{boxplots_cindex_beta0_cens_30} (30\% censoring rate) and Figure \ref{boxplots_cindex_beta0_cens_60} (60\% censoring rate). Results for the RCT data  without treatment-covariate interactions (PBC dataset) are shown in Figure \ref{boxplots_cindex_beta0_cens_30}(a) and Figure \ref{boxplots_cindex_beta0_cens_60}(a). For a censoring rate of 30\%, varying hazards, and sample sizes, the RSF based on the log-rank test splitting rule performs best. For a censoring rate of 60\%, the Cox model performs best in the nonproportional hazards setting independent of sample size, and otherwise the RSF performs best: for a total sample size of $N = 100$, the RSF using the ``maximally selected rank statistics'' splitting rule (slightly) outperforms the standard log-rank test splitting rule in the scenarios assuming a decreasing and constant hazard. Otherwise the log-rank test splitting rule gives the best results. This may indicate that the (discrimnative) performance of the RSF suffers more from higher censoring proportions in comparison to the Cox model.				
				Results for the RCT data with multiple treatment-covariate interactions (prostate cancer dataset) are shown in Figure \ref{boxplots_cindex_beta0_cens_30}(b) and Figure \ref{boxplots_cindex_beta0_cens_60}(b). For a censoring rate of 30\%, the RSF using the ``extremely randomized trees'' splitting rule performs best in the proportional hazards settings, independent of sample size. For the nonproportional hazards setting, the RSF based on the ``Harrell's C'' splitting rule performs best.   For a censoring rate of 60\%, the Cox model performs best for the higher sample sizes ($N = 200$ and $N = 400$) for two of the proportional hazards settings ($\gamma = 0.8$ and $\gamma = 1$). For all other scenarios the RSF based on the ``extremely randomized trees'' splitting rule performs best. In contrast to the scenario without treatment-covariate interactions, the RSF outperforms the Cox model with respect to discriminative performance in case of nonproportional hazards. This may indicate the ability of the RSF to handle these interactions even without prior specification in the model. \\
				The IBS estimates, which correspond to the models' overall performance (encompassing both, the models' discrimination and calibration) are shown in Figure \ref{boxplots_ibs_beta0_cens_30} (30\% censoring rate) and Figure \ref{boxplots_ibs_beta0_cens_60} (60\% censoring rate). Results for the RCT data  without treatment-covariate interactions (PBC dataset) are shown in Figure \ref{boxplots_ibs_beta0_cens_30}(a) and Figure \ref{boxplots_ibs_beta0_cens_60}(a). The Cox model has clearly the best overall performance in the nonproportional hazards settings. It also (very slightly) outperforms the RSF in the scenario with increasing hazard when either $N = 400$ for a censoring rate of 30\% or when $N \in \{200, 400\}$ for a censoring rate of 60\%. Otherwise, the RSF performs better. These differences are even more evident with decreasing total sample size. For decreasing and constant hazards,  the ``Gradient-based Brier score'' splitting rule slightly outperforms the ``log-rank test'' splitting rule for the RSF for both censoring rates and all sample sizes.
				Results for the RCT data with multiple treatment-covariate interactions (prostate cancer dataset) are shown in Figure \ref{boxplots_ibs_beta0_cens_30}(b) and Figure \ref{boxplots_ibs_beta0_cens_60}(b). For a censoring proportion of 30\%, the Cox model performs slightly better for the proportional hazards settings in case $\gamma = 0.8$ or $\gamma = 1$. In case of increasing hazards ($\gamma = 6$) or nonproportional hazards ($\gamma \in \{2,5\}$), the RSF clearly performs  better. Alternatives to the standard ``log-rank test'' splitting rule perform only slightly better, and depending on the scenario and sample size these are differing alternatives. For a censoring rate of 60\%, the Cox model performs better in some cases, especially with increasing sample size. It slightly outperforms the RSF in case of $N = 100$ when assuming an  increasing hazard ($\gamma = 6$). In case of $N = 200$, it additionally outperforms the RSF when assuming a constant hazard ($\gamma = 1$), and in case of $N = 400$, it outperforms the RSF in all proportional hazards settings ($\gamma = 0.8, \gamma = 1, \gamma = 6$). For nonproportional hazards, the RSF based on the ``extremely randomized tree'' splitting rule clearly performs best. For the remaining scenarios either the RSF using the ``Gradient-based Brier score'' ($N = 200$ and $\gamma = 0.8$) or ``Harrell's C'' splitting rule ($N = 100$ with $\gamma = 0.8$, or $\gamma = 1$) perform best. One observation is, that the Cox model has the better overall performance measured by the Integrated Brier score for the nonproportional hazards setting in the absence of treatment-covariate interactions, but performs worse in comparison to the RSF if treatment covariate interactions are present in the data, similar to the scenario with a censoring rate of 30\%. \\
				Some calibration curves at median survival time for the RCT data  without treatment-covariate interactions (PBC dataset) are shown in Figure \ref{Mayo_calib_treat_04_ph} and Figure \ref{Mayo_calib_treat_04_nonph}. Calibration curves for a proportional and nonproportional hazards setting are compared. Calibration curves for the respective scenarios for the RCT data  with multiple treatment-covariate interactions (prostate cancer dataset) are shown in Figure \ref{Byar_calib_treat_04_ph} and Figure \ref{Byar_calib_treat_04_nonph}. Additionally to the results of the Cox model and the RSF model based on the standard splitting rule, they show the results for those algorithms that outperformed these two approaches with respect to overall performance in the respective scenario. Calibration of the Cox model improves with increasing sample size while for the RSF this is at least less evident. Especially in the nonproportional hazards setting and absence of treatment-covariate interactions, the Cox model's results are better calibrated compared to the RSF (Figure \ref{Mayo_calib_treat_04_nonph}). In contrast, the difference in calibration between the two models is less obvious for the nonproportional hazards setting in case treatment-covariate interactions are present in the data (Figure \ref{Byar_calib_treat_04_nonph}). Judged by the percentiles shown as dashed lines, calibration generally varies less in the Cox model results than in the RSF results.  Deviation from perfect calibration of the RSF results is sometimes caused by a too narrow range of predictions compared to the true values, resulting in calibration curves that are too steep, most notably in the nonproportional hazards settings without treatment-covariate interactions in the data (Figure \ref{Mayo_calib_treat_04_nonph}).\\
				Computational complexity of the methods is compared in Figure \ref{comp_times}. It includes the variable selection step for the Cox model, and the grid search for finding the optimal combination of hyperparameters for the RSF.  Computational times are the lowest for the Cox model, although the RSF still has relatively low computational times for total sample sizes of $N = 100$, and even for the larger sample sizes when the RSF splitting rules ``log-rank test'',``extremely randomized trees'', or ``maximally selected rank statistics'' are used. In contrast, computational time considerably increases for larger sample sizes as well as larger number of covariates for the RSF splitting rules ``log-rank score test'', ``gradient-based Brier score'', and ``Harrell's $C$''.\\
				Complete simulation study results can be found in Supplementary Material C.1 ($C$ index), 
				C.2 (Integrated Brier score), and C.3 (calibration curves). 

				\section{Discussion and conclusions}
				An extensive neutral simulation study was performed in order to compare the performance of the Cox regression model and the RSF model for  predicting survival probabilities in RCT data. For this, we followed recommendations for neutral comparison studies \citep{Weber2019, Morris2019} to ensure an objective evaluation of the results. \\ 
				A variety of settings was considered using two publicly available RCT datasets as a reference. One dataset is characterized by the absence of treatment-covariate interactions (\citealp{UMASS}, biliary cirrhosis dataset)  and the other by two significant and one weak treatment-covariate interaction (\citealp{Byar1980}, prostate cancer dataset). In each case, different total sample sizes, values of the treatment effect, censoring rates, and properties of the hazard were considered for data simulation which may occur in other real-world datasets. Comparisons are based on measures of discrimination, calibration, and overall performance as recommended in the literature \citep{Moons2015, Steyerberg2010, McLernon2023}. \\
				Depending on the research question, different aspects of the algorithm's performance may be more important. In previous studies comparing the Cox and RSF models in real-world observational data, conclusions are usually based on the $C$ index, a measure of discrimination, but its extension and application to time-to-event medical data has been criticised \citep{Hartman2023, Vickers2010, Cook2007}. Similar to the findings of previous studies, in our simulation study the RSF predictions were usually more accurate with respect to the $C$ index, with some exceptions for the data with higher (60\%) censoring. In case of these higher censoring rates, the Cox model performed better in the nonproportional hazards setting in the absence of treatment-covariate interactions, and in case of multiple treatment-covariate interactions for constant and decreasing hazards with larger sample sizes ($N = 200, N = 400$). \\	
				With respect to overall performance measured by the Integrated Brier score, the Cox model performs considerably better in the nonproportional hazards setting for both censoring rates (30\%, 60\%) in the data without treatment-covariate interactions, but in the presence of treatment-covariate interactions, the RSF performed better for nonproportional hazards. It may be concluded that overall performance of the Cox model is only affected by deviations from the proportional hazards assumption in the presence of treatment-covariate interactions. Overall performance of the Cox model improves more visibly with increasing sample size, while RSF results are more stable across different sample sizes, maybe due to its ability for good performance even in high-dimensional settings. \\
				With respect to calibration, a measure of agreement (estimated) true and predicted outcomes, results for the RSF are worse than those for the Cox model in many cases with considerable differences. Considering overall performance, the Cox model may outperform the RSF model despite poor performance with respect to the $C$ index, due to its better calibration. It is unclear whether results may be influenced by the approach for approximating the true outcomes when estimating the calibration which is based on Cox model predictions. \\ 
				In summary, overall performance measures such as the IBS may be more suitable for drawing general conclusions about the superiority of one method over the other for predictions in time-to-event data from RCTs. Findings suggest  a poor performance of the Cox model when considering the $C$ index, a conclusion that is less obvious or even reversed when considering the IBS.	\\		 
				All currently available splitting rules for the RSF implemented in  two widely used \texttt{R} packages, \texttt{randomForestSRC} \citep{Ishwaran2021a} and \texttt{ranger} \citep{Wright2023}  were included.
				Considering the $C$ index estimates, the ``extremely randomized trees'' splitting rule most often performed better than the standard ``log-rank test'' RSF in the presence of treatment-covariate interactions. Considering the Integrated Brier score estimates, the same applies. In the absence of treatment-covariate interactions, the ``gradient-based Brier score'' splitting rule performed better than the standard RSF in scenarios with decreasing or constant hazards. Thus, it may be worthwhile, to try alternative RSF splitting rules besides the default.\\
				Additionally, computational times of some RSF splitting rules such as the standard ``log-rank test'' or the ``extremely randomized trees'' splitting rule do not extremely exceed those of the Cox model for sample sizes typically expected in RCT data in contrast to the computational time required by the RSF using other splitting rules. \\
				Results are only affected to a minor degree by the size of the treatment effect. \\
				Limitations of this simulation study are that only a limited number of datasets and scenarios, as well as a limited number of performance measures can be considered. Moreover, only  the combination of Weibull distributed survival times and uniformly distributed censoring times was considered. There also exist further RSF splitting rules \citep{Ishwaran2008} that are not currently implemented in the \texttt{R} packages \texttt{randomForestSRC} \citep{Ishwaran2021a} and \texttt{ranger} \citep{Wright2023}, so they were not included in the method comparison.\\

				\vfill
				
				\noindent \textbf{Acknowledgments}\\
				We would like to thank Prof. Dr. Sarah Friedrich, Chair for Mathematical Statistics and Artificial Intelligence in Medicine, Institute for Mathematics,
				University of Augsburg, Germany, for her support.\\
				
				
				\noindent \textbf{Funding sources}\\
				The authors are grateful for financial support of the Young Researchers Travel Scholarship Program of the University of Augsburg, and for the financial support of The International Dimension of ERASMUS+ during Ricarda Graf's research visit to the University of Reading. The sponsors had no role in study design, collection, analysis and interpretation of data, writing of the report and decision to submit the article for publication.\\
				
				\noindent \textbf{Data availability statement}\\
				The two datasets used as references for data simulations are publicly available: the RCT in primary biliary cirrhosis patients is available from a number of sources, for example from the Vanderbilt Department of Biostatistics \citep{Vanderbilt}, from the book by Fleming and Harrington (\citeyear{Fleming2005}), from kaggle \citep{kaggle}, and from the website of the University of Massachusetts \citep{UMASS}, and the RCT in prostate cancer patients is available  in the \texttt{R} package \texttt{subtee}  \citep{Ballarini2021}. The \texttt{R} code for reproducing the results of the simulation study is available on Figshare (\url{https://figshare.com/s/a4da172b22403efdaf20}).
				
				\noindent \textbf{Conflict of interest}\\
				The authors have no competing interests to declare that are relevant to the content of this article.
				
				\noindent \textbf{Ethics approval}\\
				Not applicable. 

				\bigskip

				\clearpage
				\bibliographystyle{apa}
				\bibliography{Cox_RSF_in_RCT_paper}


				
				%
				%


\end{document}